\documentclass[10pt,twocolumn,letterpaper]{article}

\usepackage{cvpr}              
\definecolor{cvprblue}{rgb}{0.21,0.49,0.74}
\usepackage[pagebackref,breaklinks,colorlinks,allcolors=cvprblue]{hyperref}
\usepackage{adjustbox}  
\usepackage{siunitx}    
\usepackage{booktabs}   
\usepackage{makecell}
\usepackage{multirow}
\usepackage{placeins}
\usepackage{tablefootnote}
\usepackage{algorithm}
\usepackage{algpseudocode}
\DeclareMathOperator*{\rowmax}{rowmax}
\DeclareMathOperator*{\rowsum}{rowsum}
\DeclareMathOperator*{\TopK}{TopK}

\def\confName{CVPR}
\def\confYear{2026}

\title{Speed3R: Sparse Feed-forward 3D Reconstruction Models}

\author{
    Weining Ren$^{1}$ \quad 
    Xiao Tan$^{2}$ \quad 
    Kai Han$^{1,}$\thanks{Corresponding author.} \\
    $^{1}$The University of Hong Kong \qquad $^{2}$Baidu AMU \\
    \href{https://visual-ai.github.io/speed3r/}{https://visual-ai.github.io/speed3r/}
}

\begin{document}
\maketitle
\begin{abstract}
While recent feed-forward 3D reconstruction models accelerate 3D reconstruction by jointly inferring dense geometry and camera poses in a single pass, their reliance on dense attention imposes a quadratic complexity, creating a prohibitive computational bottleneck that severely limits inference speed. To resolve this, we introduce Speed3R, an end-to-end trainable model inspired by the core principle of Structure-from-Motion: that a sparse set of keypoints is sufficient for robust pose estimation. Speed3R features a dual-branch attention mechanism where a compression branch creates a coarse contextual prior to guide a selection branch, which performs fine-grained attention only on the most informative image tokens. This strategy mimics the efficiency of traditional keypoint matching, achieving a remarkable 12.4x inference speedup on 1000-view sequences, while introducing a minimal, controlled trade-off in geometric accuracy. Validated on standard benchmarks with both VGGT and $\pi^3$ backbones, our method delivers high-quality reconstructions at a fraction of computational cost, paving the way for efficient large-scale scene modeling. 
\end{abstract}    
\section{Introduction}
\label{sec:intro}

\begin{figure}[t]
\centering
    \includegraphics[width=1.0\linewidth]{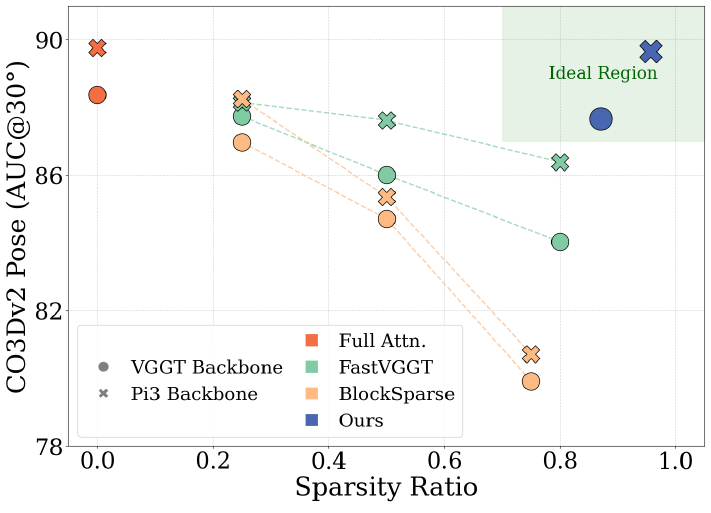}
    \caption{\textbf{Performance comparison of different methods}. CO3Dv2 pose estimation accuracy vs. sparsity ratio, highlighting the trade-off between sparsity and accuracy. The Ideal Region marks the desired balance of high accuracy and high sparsity.}
    \label{fig:intro}
\end{figure}

Classical 3D reconstruction methods are fundamentally rooted in the principle of sparsity. They traditionally begin with the detection and matching of a sparse set of salient keypoints. Early methods relied on handcrafted descriptors~\citep{lowe1999object, bay2008speeded, rublee2011orb,detone2018superpoint, sarlin2020superglue} to establish 2D correspondences across views. Initial geometric relationships are estimated from these matches, followed by a large-scale, iterative optimization Bundle Adjustment~\cite{hartley_multiple_2000}, which refines both the camera parameters and the sparse 3D structure. This multi-stage process, exemplified by highly successful and robust systems~\citep{schoenberger2016sfm, schoenberger2016mvs, pan2024glomap}, demonstrates a core insight: a sparse, carefully selected set of 2D features is sufficient to establish robust geometric constraints and reconstruct an accurate sparse point cloud.

Recently, the advent of feed-forward models~\cite{wang2024dust3r, yang2025fast3r,wang2025vggt, wang2025pi3} has revolutionized 3D reconstruction, enabling the joint inference of dense geometry and camera poses from multiple views in a single network pass. These methods, often built upon powerful vision transformer architectures~\cite{oquab2023dinov2, dosovitskiy2020vit}, bypass the complex, multi-stage pipelines of classical approaches like Structure-from-Motion (SfM) and Multi-View Stereo (MVS). However, this progress comes at a price. The dense global attention along all image tokens imposes a quadratic complexity with respect to the number of input image tokens. This creates a prohibitive computational bottleneck that severely limits inference speed, making the processing of a large number of views or high-resolution images intractable.

To address the computational bottleneck, we propose Speed3R, an end-to-end trainable model that accelerates inference while retaining the benefits of feed-forward reconstruction. Our approach integrates two key inspirations: the classical Structure-from-Motion (SfM) principle that robust geometric estimation relies on sparse keypoints rather than dense pixel comparisons, and the success of sparse attention in Large Language Models~\cite{yuan2025native, lu2025moba, gao2024seerattention} and Video Diffusion Models~\cite{zhang2025vsa, cai2025moc}. Speed3R operationalizes these insights through a dual-branch attention mechanism. A compression branch generates a global scene summary, guiding a selection branch that performs fine-grained attention on a small subset of informative tokens. This design emulates traditional keypoint-based methods, concentrating computation where it is most impactful, and achieves significant efficiency gains without sacrificing accuracy.

Through this tailored sparse attention mechanism, Speed3R achieves a remarkable 12.4x inference speedup on 1000-view sequences. Our extensive experiments demonstrate that this substantial acceleration is achieved while introducing only a minimal and controlled trade-off in geometric accuracy, establishing a new Pareto-optimal frontier in the efficiency-fidelity landscape. We validate the robustness and generalizability of our method by integrating it with state-of-the-art backbones, including VGGT~\cite{wang2025vggt} and $\pi^3$~\cite{wang2025pi3}, and demonstrate consistently superior performance on standard benchmarks than training-free methods. With zero-shot test-time adaptation, our method can outperform dense models on long sequences. By rethinking the attention mechanism for the global attention layer, Speed3R delivers high-quality reconstructions at a fraction of the computational cost, paving the way for practical and efficient large-scale scene modeling. Overall, we make the following contributions:
\begin{itemize}
    \item We propose Speed3R, a novel dual-branch feed-forward reconstruction models with trainable sparse attention mechanism that mimics classical SfM by focusing computation on a small, informative subset of tokens.
    \item We achieve a new SoTA in the efficiency-accuracy trade-off, demonstrating a 12.4x speedup for a 1000-view sequence with a minimal impact on geometric accuracy.
    \item We validate the robustness and generalizability of Speed3R, showing it integrates with various backbones and outperforms competing training-free methods.
\end{itemize}

\section{Related Works}
\paragraph{Optimization-based Multi-view Reconstruction.}
Traditional 3D reconstruction methods operate on a sparse-to-dense paradigm. The process begins with Structure-from-Motion (SfM)~\cite{schoenberger2016sfm, pan2024glomap}, where a sparse set of matched keypoints~\cite{lowe1999object, bay2008speeded, rublee2011orb} is used to optimize camera poses and a sparse point cloud via bundle adjustment~\cite{hartley_multiple_2000}. This sparse geometric backbone is then densified using Multi-View Stereo (MVS) algorithms~\cite{schoenberger2016mvs}. While deep learning has modernized this pipeline with learned feature matchers~\cite{sarlin2020superglue,lindenberger2023lightglue}, learned MVS cost volumes~\cite{yao2018mvsnet,geomvsnet}, and differentiable optimization~\cite{teed2021droid,wang2024vggsfm}, the core methodology remains dependent on an initial sparse representation and iterative optimization, making it inherently slow and computationally demanding.

\paragraph{Feed-forward 3D Reconstruction.}
A recent family of methods enables the joint inference of camera poses and dense geometry in a single forward pass. This paradigm, pioneered by DUSt3R~\cite{wang2024dust3r} for pairwise input, was quickly refined with dedicated feature heads for improved matching (MASt3R~\cite{mast3r_eccv24}) and extended to handle sequential inputs (CUT3R~\cite{cut3r}, Spann3R~\cite{wang20243d}). Architectural innovations, such as the elegant and effective design of VGGT~\cite{wang2025vggt} and the permutation-equivariant structure of $\pi^3$~\cite{wang2025pi3}, have pushed the state of the art. However, the reliance of these advanced models on dense, all-to-all attention for global information exchange creates a significant computational bottleneck, especially for long sequences. To mitigate this, several training-free sparsification approaches have been proposed. For instance, FastVGGT~\cite{shen2025fastvggt} employs a token merge-and-unmerge strategy, while Block Sparse VGGT~\cite{wang2025faster} applies top-k attention. Because these methods are not training-aware, their ability to sparsify the model is limited; aggressive pruning results in a notable degradation of reconstruction accuracy.

\paragraph{Sparse Attention.}
To mitigate the quadratic complexity of standard attention, various sparse attention methods have been proposed. Early approaches employed fixed, data-agnostic patterns like local windows (StreamingLLM~\cite{xiao2023streamingllm}) or dilated windows (Longformer~\cite{Beltagy2020Longformer}). More advanced methods are dynamic, such as those that prune the KV cache during inference (H2O~\cite{zhang2023h2o}), although these post-hoc optimizations do not accelerate the training phase. Another class of dynamic methods performs query-aware token selection using techniques like clustering~\cite{liu2025clusterkv} or heuristic scoring~\cite{tang2024quest}; however, these often suffer from non-differentiable operations or non-contiguous memory access, hindering end-to-end training. To resolve these issues, recent works NSA~\cite{yuan2025native} and MOBA~\cite{lu2025moba} present a breakthrough with their trainable sparse attention and show comparable results with full attention. The success of these methods has led to their application in various domains, including video generation~\cite{zhang2025vsa, cai2025moc} and 3D generation~\cite{wu2025direct3ds2, he2025triposf}. Its core philosophy of adaptively selecting important tokens is particularly well-suited to the sparse nature of 3D reconstruction. Highly inspired by NSA, our work aims to develop an efficient sparse attention mechanism tailored for 3D feed-forward reconstruction.
\section{Method}

\begin{figure*}[t]
    \centering
    \includegraphics[width=\linewidth]{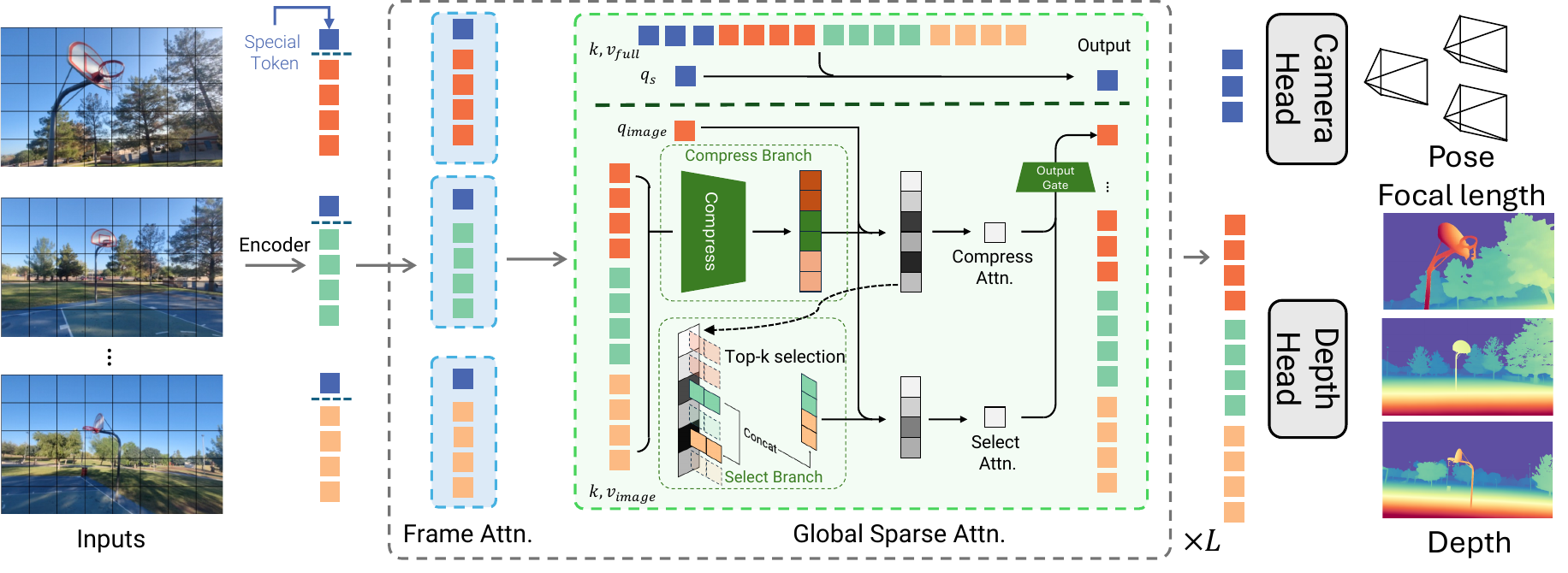}
    \caption{\textbf{Architecture.} Our model processes a sequence of input images through a shared feature encoder. The resulting tokens are then processed by a series of transformer blocks that alternate between local \textbf{Frame Attention} (within each view) and our proposed \textbf{Global Sparse Attention} (across all views). The GSA module efficiently integrates global information by decomposing attention into a \textbf{Compression Branch} for coarse context and a \textbf{Selection Branch} for fine-grained details, guided by a Top-k selection mechanism. Finally, updated tokens are passed to task-specific heads to predict camera pose and dense depth maps.}
    \label{fig:arch}
\end{figure*}

Our primary goal is to develop a feed-forward 3D reconstruction model capable of efficiently processing a large number of views. To address the scalability bottleneck, we introduce a novel attention mechanism, \textbf{Global Sparse Attention (GSA)}, designed as a drop-in replacement for the original global attention module.

\subsection{Architecture Overview}
We adopt the foundational architecture of recent feed-forward reconstruction models, illustrated in Figure~\ref{fig:arch}. The model processes a sequence of input images to jointly predict camera parameters (pose and focal length) and dense depth maps. The architecture consists of three main stages:
\begin{itemize}
    \item \textbf{Per-frame Feature Encoder:} A sequence of N images \(\{I_i\}_{i=1}^N\) are independently passed through vision encoder (e.g., DINOv2~\cite{oquab2023dinov2}) to extract patch-based feature tokens.
    \item \textbf{Alternating Attention Transformer:} A series of transformer blocks process the tokens. These blocks alternate between local Attention, which operates on tokens within a single frame, and our proposed Global Sparse Attention, which efficiently fuses information across all frames.
    \item \textbf{Task-Specific Prediction Heads:} The refined tokens are used by downstream heads to predict per-view camera parameters \(\{\hat{C_i}\}_{i=1}^N\), depth maps \(\{\hat{D_i}\}_{i=1}^N\) and their associated uncertainty \(\{\hat{\alpha_i}\}_{i=1}^N\).
\end{itemize}
Our key innovation lies in the second stage, where we replace the computationally intensive global full-attention layer with our GSA module. For all other components, including the feature encoder, frame attention, and prediction heads, we follow the design of the original base models (VGGT~\cite{wang2025vggt} or $\pi^3$~\cite{wang2025pi3}).

\subsection{Global Sparse Attention (GSA)}
The GSA module is designed to approximate full attention with significantly reduced computational complexity, making it scalable to a large number of views. Its core principle is to leverage a coarse, low-resolution representation of the scene, computed by the \textbf{Compression Branch}, to guide the selection of a sparse subset of high-resolution tokens for the \textbf{Selection Branch}. This coarse-to-fine strategy enables the model to efficiently build a global understanding of the scene while focusing its limited computational budget on the most salient, keypoint region details.

Let the input to the GSA block be a sequence of tokens $X \in \mathbb{R}^{M \times C}$ (batch dimension is omitted), which is a concatenation of special tokens $X_{\text{spec}} \in \mathbb{R}^{M_{\text{spec}} \times C}$ and image tokens $X_{\text{img}} \in \mathbb{R}^{M_{\text{img}} \times C}$, where $M = M_{\text{spec}} + M_{\text{img}}$. We first project the entire sequence into queries, keys, and values with dimension $d$ using linear transformations $W_Q, W_K, W_V$. The resulting tensors are then partitioned corresponding to their original token types:
\begin{equation}
    Q = \begin{bmatrix} Q_{\text{spec}} \\ Q_{\text{img}} \end{bmatrix}, \quad K = \begin{bmatrix} K_{\text{spec}} \\ K_{\text{img}} \end{bmatrix}, \quad V = \begin{bmatrix} V_{\text{spec}} \\ V_{\text{img}} \end{bmatrix}
\end{equation}
The attention computation then proceeds differently for the two token types.

\paragraph{Full Attention for Special Tokens.}
Special tokens serve as global information bottlenecks and are critical for tasks like pose estimation. To ensure they have a comprehensive view of the entire scene, they perform standard, dense self-attention over all other tokens. The output for these tokens, $O_{\text{spec}}$, is computed as:
\begin{equation}
    O_{\text{spec}} = \text{Attention}(Q_{\text{spec}}, K, V) = \text{softmax}\left(\frac{Q_{\text{spec}} K^T}{\sqrt{d_k}}\right) V
\end{equation}
This operation, while quadratic, is inexpensive as the number of special tokens ($M_{\text{spec}}$) is very small.


\paragraph{Sparse Attention for Image Tokens.}
For the vast number of image tokens, we employ a two-branch strategy that computes a coarse global summary and fine-grained local details in sequential. The outputs from these branches are then dynamically fused using a learned gating mechanism, allowing the model to adaptively balance global context against local specificity for each token.

\paragraph{Compression Branch.}
This branch provides a coarse but comprehensive summary of the scene in a highly efficient manner. To create a computationally inexpensive proxy for global attention, we spatially downsample QKV tensors ($Q_{\text{img}}$, $K_{\text{img}}$, $V_{\text{img}}$). This is achieved using a non-overlapping average pooling operation with a window size of $s \times s$, yielding compressed tensors $Q_{\text{comp}}$, $K_{\text{comp}}$, and $V_{\text{comp}}$, all of size $\mathbb{R}^{M'_{\text{img}} \times d}$, where $M'_{\text{img}} = M_{\text{img}} / s^2$.

The attention calculation of this branch is then performed entirely within this compressed space, producing a coarse output tensor $O'_{\text{comp}}$. Additionally, to guide the subsequent selection process, a score matrix $S_{\text{guide}}$ is computed from the compressed queries and keys. The coarse output $O'_{\text{comp}}$ is subsequently upsampled to the original image token resolution using nearest-neighbor interpolation, which assigns the same context vector to all fine-grained tokens that belong to the same spatial window. This yields the final branch output, $O_{\text{comp}}$.

\begin{align}
    O'_{\text{comp}} &= \text{Attention}(Q_{\text{comp}}, K_{\text{comp}}, V_{\text{comp}}) \in \mathbb{R}^{M'_{\text{img}} \times d} \\
    S_{\text{guide}} &= Q_{\text{comp}} K_{\text{comp}}^T \in \mathbb{R}^{M'_{\text{img}} \times M'_{\text{img}}} \\
    O_{\text{comp}} &= \text{Upsample}(O'_{\text{comp}}) \in \mathbb{R}^{M_{\text{img}} \times d}
\end{align}

\paragraph{Selection Branch.}
To recover fine-grained attention, this branch performs attention on a small subset of the original, full-resolution key-value pairs. To guide the selection, we use the pre-computed relevance score matrix $S_{\text{guide}}$ to identify the most relevant coarse regions for each query. For each query, we use a $\text{TopKSelect}(\cdot)$ function on $S_{\text{guide}}$ to identify the indices of the most relevant \textit{regions}. Queries belonging to the same compression window share the same set of KV pairs. The original, full-resolution key-value pairs ($K_{\text{img}}, V_{\text{img}}$) corresponding to these top regions are then selected, forming the sparse sets $K_{\text{sel}}$ and $V_{\text{sel}}$. The fine-grained output $O_{\text{sel}}$ is computed by attending only to this small subset:

\begin{equation}
    O_{\text{sel}} = \text{Attention}(Q_{\text{img}}, K_{\text{sel}}, V_{\text{sel}})
\end{equation}
This operation is highly efficient as each query only attends to $k \ll M_{\text{img}}$ tokens. 
\paragraph{Gated Aggregation.}
The outputs from the two branches are combined using a learnable gating mechanism that weights their contributions based on the query itself. A gating vector $g \in \mathbb{R}^{M_{\text{img}} \times d}$ is computed from the image queries, and the final output for the image tokens, $O_{\text{img}}$, is a dynamic, weighted sum:
\begin{align}
    g &= \sigma(W_gQ_{\text{img}}) \\
    O_{\text{img}} &= g \odot O_{\text{comp}} + (1 - g) \odot O_{\text{sel}}
\end{align}
where $\sigma$ is the sigmoid function, $W_g$ is a learned projection matrix, and $\odot$ denotes element-wise multiplication. This allows the model to decide for each token whether to rely more on the global summary from the Compression Branch or the specific details from the Selection Branch. 

\paragraph{Final Output.}
The final output of the GSA layer, $O_{\text{GSA}}$, is produced by concatenating the outputs from the two pathways in their original order:
\begin{equation}
    O_{\text{GSA}} = \text{concat}(O_{\text{spec}}, O_{\text{img}}) \in \mathbb{R}^{M \times d}
\end{equation}

\paragraph{Efficient Kernel Implementation.}
A naive implementation of the Compression Branch with $\text{TopKSelect}(\cdot)$ is inefficient, bottlenecked by the large memory footprint of the full score matrix  $S_{\text{guide}}$. To overcome this, we developed a fused GSA kernel in Triton~\cite{tillet2019triton}. Our approach integrates a streaming Top-K algorithm directly into the FlashAttention~\cite{dao2023flashattention2} workflow. As our kernel computes score matrix tiles in fast on-chip SRAM, it not only performs the online softmax but simultaneously maintains a running set of the top-k indices and scores for each query. This allows the selection of the most relevant keys and the calculation of compression output to occur in a single, fused pass over the input data. By doing so, we avoid materializing the full score matrix and maximize data locality. Implementation details are provided in the supplementary material.

\subsection{Speed3R-VGGT}
\label{sec:Speed3R-vggt}
We instantiate our method on the VGGT architecture, terming this variant \textbf{Speed3R-VGGT}. The VGGT model architecture is unique in that it designates the first frame of a sequence as a global reference and utilizes dedicated camera tokens to encode pose information. 

To ensure this critical global reference is not lost by the sparse attention mechanism, we adapt the GSA's Selection Branch. For any given query, its attention set $(K_{\text{sel}}, V_{\text{sel}})$ is constructed from the \textbf{concatenation} of two groups:
    (i) A fixed global context set comprising all tokens from the reference frame and frames sampled at 100-frame intervals.
    (ii) The dynamically selected Top-K image tokens windows from non-reference frames, identified by our standard selection process. 
This hybrid approach guarantees that while the model can efficiently focus on salient local details in subsequent frames, it never loses sight of the foundational reference frame and camera parameters.

To transfer the performance of the original dense model to our efficient sparse variant, we employ a knowledge distillation strategy. The student model (Speed3R-VGGT) is trained to replicate the outputs of its pre-trained dense teacher. The teacher's predictions for depth and camera pose serve as the pseudo ground truth for the student. The total training loss is a weighted sum of a depth distillation loss and a camera pose distillation loss:
\begin{equation}
    \mathcal{L}_{\text{total}} = \mathcal{L}_{\text{depth}} + \lambda \mathcal{L}_{\text{camera}}
\end{equation}
where the losses are inherent from original VGGT losses.

\subsection{Speed3R-$\boldsymbol{\pi^3}$}
Similarly, we apply our method to the $\pi^3$~\cite{wang2025pi3} architecture to create Speed3R-$\boldsymbol{\pi^3}$. Unlike VGGT~\cite{wang2025vggt}, $\pi^3$~\cite{wang2025pi3} does not rely on reference frame or camera tokens, which allows for a more direct application of our GSA module as described in Section 3.2. Additionally, we observe empirically that the register tokens used in the original $\pi^3$ can be omitted in our sparse variant without performance drop, further simplifying the model without special tokens. We follow the same distillation strategy, using the original dense $\pi^3$ model as the teacher to train our Speed3R-$\pi^3$ student with relative pose loss and depth loss. The total loss function is identical in form to that used for original $\pi^3$.

\subsection{Training Details}
We train our model on a mixture of seven datasets: ArkitScene~\cite{dehghan2021arkitscenes}, Scannet++~\cite{yeshwanthliu2023scannetpp}, DL3DV~\cite{ling2024dl3dv}, CO3D~\cite{reizenstein21co3d}, Hypersim~\cite{hypersim}, WildRGBD~\cite{xia2024rgbd}, and VirtualKitti2~\cite{cabon2020virtual}. Some datasets were downsampled for storage efficiency (see supplementary material), and the sparse model's weights were initialized from its pre-trained dense counterpart. The model was trained for 80 epochs (800 steps per epoch) over approximately 7 days on 8 NVIDIA H20 GPUs. We followed most original dense model's setting, but with two adjustments: the learning rate was initialized to $1 \times 10^{-5}$, and a gradient accumulation factor of 4 was used to achieve an effective batch size of 32.

\section{Experiments}

We evaluate our sparse models against their dense counterparts and two training-free baselines: FastVGGT~\cite{shen2025fastvggt} and Block-Sparse VGGT~\cite{wang2025faster}. Variants of these baselines with VGGT/$\pi^3$ are referred to as FastVGGT-VGGT/$\pi^3$ and Block-Sparse VGGT/$\pi^3$, respectively. Unless stated otherwise, we use following parameters.  Our method employs a 4x4 compression window and selects the top-32 blocks for selective attention. For the baselines, we adopt their default configurations: a 0.9 merge ratio for FastVGGT~\cite{shen2025fastvggt} and a 0.75 sparsity ratio for Block-Sparse VGGT/$\pi^3$~\cite{wang2025faster}. All inference times are benchmarked on a single H100 GPU.

\subsection{Two-view Pose Estimation}
\begin{table}[t!]
\centering
\caption{
\textbf{Pair-wise pose results on ScanNet-1500~\citep{dai2017scannet,sarlin2020superglue}.} We report the Area Under the Curve (AUC) of the pose error at different thresholds. Best results per backbone are marked in \textbf{bold}.}
\label{tab:relpose_scannet}
\adjustbox{valign=t,width=\linewidth}{
\begin{tabular}{lccc}
\toprule
\multirow{2}{*}{Methods} & \multicolumn{3}{c}{ScanNet1500}  \\

   & AUC@5 $\uparrow$ & AUC@10 $\uparrow$ & AUC@20 $\uparrow$ \\
\midrule

VGGT~\citep{wang2025vggt} & \textcolor{gray}{37.45} & \textcolor{gray}{59.24} & \textcolor{gray}{75.69} \\
Block Sparse-VGGT~\cite{wang2025faster} & 33.21 & 55.11 & 72.51 \\
FastVGGT-VGGT~\cite{shen2025fastvggt} & 33.59 & 56.21 & 73.47 \\
Speed3R-VGGT & \textbf{37.02} & \textbf{59.11} & \textbf{75.62} \\

\cmidrule{1-4}

$\pi^3$~\cite{wang2025pi3} & \textcolor{gray}{\textbf{38.76}} & \textcolor{gray}{\textbf{61.57}} & \textcolor{gray}{\textbf{77.61}} \\
Block Sparse-$\pi^3$~\cite{wang2025faster} & 35.13 & 57.74 & 74.98 \\
FastVGGT-$\pi^3$~\cite{shen2025fastvggt} & 34.87 & 58.31 & 75.51 \\
Speed3R-$\pi^3$  & \textbf{36.97} & \textbf{59.83} & \textbf{76.38} \\
\bottomrule
\end{tabular}
}
\end{table}
We first evaluate our method on the ScanNet relative pose estimation task~\cite{dai2017scannet, sarlin2020superglue}, reporting the Area Under the Curve (AUC) of the pose error. This metric is the area under an accuracy-threshold curve where per-pair accuracy is defined as the minimum of the Relative Rotation Accuracy (RRA) and Relative Translation Accuracy (RTA) at different thresholds. The results in Table~\ref{tab:relpose_scannet} show that our method (Speed3R) consistently outperforms competing training-free sparse methods. Given that this benchmark evaluates performance under large viewpoint changes, this result highlights our method's superior robustness to such variations. Notably, Speed3R-VGGT also achieves performance nearly on par with the original dense VGGT model, and this advantage holds for the more powerful $\pi^3$ backbone, confirming the general effectiveness of our approach.

\subsection{Multi-view Pose Estimation}
\begin{table}[!tp]
\begin{center}
\caption{\textbf{Pose Estimation on RE10k~\citep{zhou2018stereo} and CO3Dv2~\citep{reizenstein21co3d}.} These datasets contain an average of 10 images per scene. }
\label{tab:re10kco3d}
\adjustbox{valign=t,width=\linewidth}{
\sisetup{detect-all=true,detect-weight=true}
\begin{tabular}{lccc}
\noalign{\hrule height 1.2pt}  
Method & \makecell{Sparse Ratio (\%)/ \\ Anchor Number } & \makecell{RE10K \\ AUC@30 $\uparrow$} & \makecell{CO3Dv2 \\ AUC@30 $\uparrow$} \\
\midrule

VGGT~\citep{wang2025vggt} & 0 &  \textcolor{gray}{74.17} & \textcolor{gray}{88.33}  \\
\hline
Block Sparse-VGGT~\citep{wang2025faster} & 25 & 71.79 & 86.98  \\
                                         & 50 & 68.25 & 84.71  \\
                                         & 75 & 63.82 & 79.92  \\

\hline

 SAIL-Recon~\citep{deng2025sail}  & 10 anchor & 74.31 & 87.63 \\
                                  & 5 anchor  & 72.66 & 84.25 \\
                                  & 2 anchor  & 69.11 & 80.03 \\
 \hline

 FastVGGT~\citep{shen2025fastvggt}  & 25 & 72.97 & 87.74  \\
                                    & 50 & 71.55 & 86.01  \\
                                    & 82 & 69.99 & 84.03  \\

  \hline

Speed3R-VGGT                      & 84 & 74.81 & 87.71  \\

\noalign{\hrule height 1.2pt}  
$\pi^3$~\citep{wang2025pi3}                 & 0 & \textcolor{gray}{87.37} & \textcolor{gray}{89.67} \\
\hline
Block Sparse-$\pi^3$~\citep{wang2025faster} & 25 & 85.18 & 88.25 \\
                                        & 50 & 81.29 & 85.36 \\
                                        & 75 & 75.39 & 80.72 \\

\hline
FastVGGT-$\pi^3$~\citep{shen2025fastvggt} & 25 & 87.26 & 88.15 \\
                                      & 50 & 86.67 & 87.62 \\
                                      & 90 & 86.04 & 86.39 \\
\hline
Speed3R-$\pi^3$                           & 94 & 87.17 & 89.41 \\
\noalign{\hrule height 1.2pt}  
\end{tabular}
}
\end{center}
\end{table}
We further validate our approach on the multi-view RE10k~\citep{zhou2018stereo} and CO3Dv2~\citep{reizenstein21co3d} benchmarks, which include 10 frames per scene, with comparisons to SAIL-Recon~\cite{deng2025sail}. As shown in Table~\ref{tab:re10kco3d} and visualized in Figure~\ref{fig:intro}, Speed3R establishes a new Pareto-optimal frontier for accuracy and efficiency. It consistently outperforms all competing sparse and anchor-based methods, often at significantly higher sparsity ratios. This superiority is demonstrated by two key results: first, our Speed3R-VGGT model (84\% sparsity) surpasses the dense VGGT baseline on RE10k; second, the Speed3R-$\pi^3$ model (94\% sparsity) nearly matches the performance of its dense counterpart. These findings demonstrate that Speed3R effectively prunes redundant computations while preserving performance-critical information.

\subsection{Long-sequence Pose Estimation}
\begin{table}[!tp]
\begin{center}
\caption{\textbf{Pose Estimation on Tanks \& Temples~\citep{Knapitsch2017}.} This dataset contains, an average of 300 images per scene. Best results and second best results are \textbf{in bold} and \underline{underlined} separately.}
\label{tab:tnt_pose}
\adjustbox{valign=t,width=\linewidth}{
\begin{tabular}{lcccc}
\toprule
 Method  & RRA@5 $\uparrow$ & RTA@5 $\uparrow$ & AUC@30 $\uparrow$ & Time [s] $\downarrow$ \\
\midrule

VGGT~\citep{wang2025vggt} & \textcolor{gray}{70.29} & \textcolor{gray}{79.30} & \textcolor{gray}{77.67} & \textcolor{gray}{34.51} \\
Block Sparse-VGGT~\citep{wang2025faster} & 66.83 & 71.29 & 74.15  & \underline{10.79} \\
 SAIL-Recon(20 anchor)~\citep{deng2025sail}  & 68.34 & 73.77 & 74.98 & 20.35 \\
 SAIL-Recon(100 anchor)~\citep{deng2025sail}  & \textbf{69.72} & 75.16 & \underline{75.70} & 53.02 \\
 FastVGGT~\citep{shen2025fastvggt}  & 69.28 & \textbf{77.98} & 76.29  & 15.98 \\
 Speed3R-VGGT & \underline{69.51} & \underline{77.81} & \textbf{76.57}  & \textbf{6.55} \\

\cmidrule{1-5}
$\pi^3$~\citep{wang2025pi3} & \textcolor{gray}{72.14} & \textcolor{gray}{81.26} & \textcolor{gray}{79.63} & \textcolor{gray}{22.32} \\
Block Sparse-$\pi^3$~\citep{wang2025faster} & 67.85 & 78.91 & 76.64  & \underline{8.16} \\
FastVGGT-$\pi^3$~\citep{shen2025fastvggt} & \underline{69.78} & \underline{79.51} & \underline{77.76} & 11.96 \\
Speed3R-$\pi^3$  & \textbf{70.72} & \textbf{80.72} & \textbf{79.77} & \textbf{4.19} \\
\bottomrule

\end{tabular}
}
\end{center}
\end{table} 
We further evaluate on the large-scale Tanks \& Temples (T\&T) benchmark~\citep{Knapitsch2017} with an average of 300 images per sequence. As shown in Table~\ref{tab:tnt_pose}, Speed3R demonstrates a state-of-the-art balance of accuracy and speed. With the VGGT backbone, our method is by far the fastest (6.65s), achieving a 5.2x speedup over the dense baseline while maintaining top-tier accuracy, including the best AUC@30 score among all sparse methods. The same holds for the $\pi^3$ backbone. Speed3R-$\pi^3$ simultaneously achieving the highest accuracy across all metrics and the lowest runtime (4.19s) among all sparse methods. It nearly matches the performance of the dense $\pi^3$ model while being 5.3x faster. These results confirm that Speed3R excels at Pareto-optimal that are both highly accurate and exceptionally efficient for large-scale feed-forward pose estimation.

\subsection{Pointmap Estimation}
\begin{table*}[ht]
  \centering
  \caption{\textbf{Pointmap Estimation on the DTU~\cite{dtudataset} and ETH3D~\cite{eth3d} datasets}. The arrows ($\downarrow$/$\uparrow$) indicate whether lower or higher values are better. Best results are highlighted in \textbf{bold}.}
  \label{tab:comparison_results}
  \resizebox{\linewidth}{!}{%
  \begin{tabular}{l cc cc cc cc cc cc}
    \toprule
    \textbf{Method} & \multicolumn{6}{c}{\textbf{DTU}~\cite{dtudataset}} & \multicolumn{6}{c}{\textbf{ETH3D}~\cite{eth3d}} \\
    \cmidrule(lr){2-7} \cmidrule(lr){8-13}
    & \multicolumn{2}{c}{Acc. $\downarrow$} & \multicolumn{2}{c}{Comp. $\downarrow$} & \multicolumn{2}{c}{N.C. $\uparrow$} & \multicolumn{2}{c}{Acc. $\downarrow$} & \multicolumn{2}{c}{Comp. $\downarrow$} & \multicolumn{2}{c}{N.C. $\uparrow$} \\
    & Mean & Med. & Mean & Med. & Mean & Med. & Mean & Med. & Mean & Med. & Mean & Med. \\
    \midrule

    VGGT~\cite{wang2025vggt}         & \textcolor{gray}{1.403} & \textcolor{gray}{0.802} & \textcolor{gray}{2.566} & \textcolor{gray}{1.307} & \textcolor{gray}{0.658} & \textcolor{gray}{0.742} & \textcolor{gray}{0.289} & \textcolor{gray}{0.192} & \textcolor{gray}{0.294} & \textcolor{gray}{0.173} & \textcolor{gray}{0.847} & \textcolor{gray}{0.953}       \\
        
    Block Sparse-VGGT~\cite{wang2025faster}   & 1.966 &  1.052 &    2.311       &    1.135   & 0.647 & 0.715 & 0.861 & 0.754 & 1.171 & 0.812 & 0.681 & 0.772\\

    FastVGGT-VGGT~\cite{shen2025fastvggt}   & 1.466 &  \textbf{0.786} &    2.385       &    1.188 & 0.654 & 0.736 & 0.510 & 0.379 & 0.580 & 0.354 & 0.788 & 0.913\\
    
    Speed3R-VGGT   & \textbf{1.426} &  0.827 & \textbf{2.179} &    \textbf{1.101} & \textbf{0.657} & \textbf{0.740} & \textbf{0.295} & \textbf{0.190} & \textbf{0.289} & \textbf{0.168} & \textbf{0.853} & \textbf{0.953} \\
    
    \midrule
    $\pi^3$~\cite{wang2025pi3}         & \textcolor{gray}{1.151} & \textcolor{gray}{0.622} & \textcolor{gray}{1.793} & \textcolor{gray}{0.629} & \textcolor{gray}{0.668} & \textcolor{gray}{0.754} & \textcolor{gray}{0.194} & \textcolor{gray}{0.130} & \textcolor{gray}{0.220} & \textcolor{gray}{0.135} & \textcolor{gray}{0.867} & \textcolor{gray}{0.965} \\

    Block Sparse-$\pi^3$~\cite{wang2025faster}   & 2.434 &  1.130 &    2.714       &    1.004     & \textbf{0.664} & \textbf{0.749} & 0.313 & 0.235 & 0.439 & 0.276 & 0.816 & 0.951\\

    FastVGGT-$\pi^3$~\cite{shen2025fastvggt}         & 1.255          & 0.737          & 2.250          & 0.857          & 0.650          & 0.730          & 0.291          & 0.215          & 0.291          & 0.179          & 0.841          & 0.961          \\

    Speed3R-$\pi^3$      & \textbf{1.175}          & \textbf{0.710}          & \textbf{2.037}         &\textbf{0.731}         & 0.657        & 0.739         & \textbf{0.198}          & \textbf{0.136}        & \textbf{0.213}          & \textbf{0.126}          & \textbf{0.878}          & \textbf{0.970}          \\
    \bottomrule
  \end{tabular}%
  }
  \label{tab:pi3_comp}
\end{table*}
\begin{figure*}[!htbp] 
    \centering

    \hspace{1em}
    \begin{subfigure}{0.16\textwidth}
        \subcaption*{Input Image}
    \end{subfigure}\hfill
    \begin{subfigure}{0.16\textwidth}
        \subcaption*{Ground Truth}
    \end{subfigure}\hfill
    \begin{subfigure}{0.16\textwidth}
        \subcaption*{Dense Model~\cite{wang2025vggt, wang2025pi3}}
    \end{subfigure}\hfill
    \begin{subfigure}{0.16\textwidth}
        \subcaption*{Block Sparse~\cite{wang2025faster}}
    \end{subfigure}\hfill
    \begin{subfigure}{0.16\textwidth}
        \subcaption*{FastVGGT~\cite{shen2025fastvggt}}
    \end{subfigure}\hfill
    \begin{subfigure}{0.16\textwidth}
        \subcaption*{Ours}
    \end{subfigure}

    \rotatebox{90}{\makebox[4em][c]{VGGT}}%
    \hspace{0.5em}
    \begin{subfigure}{0.16\textwidth}
        \includegraphics[width=\linewidth]{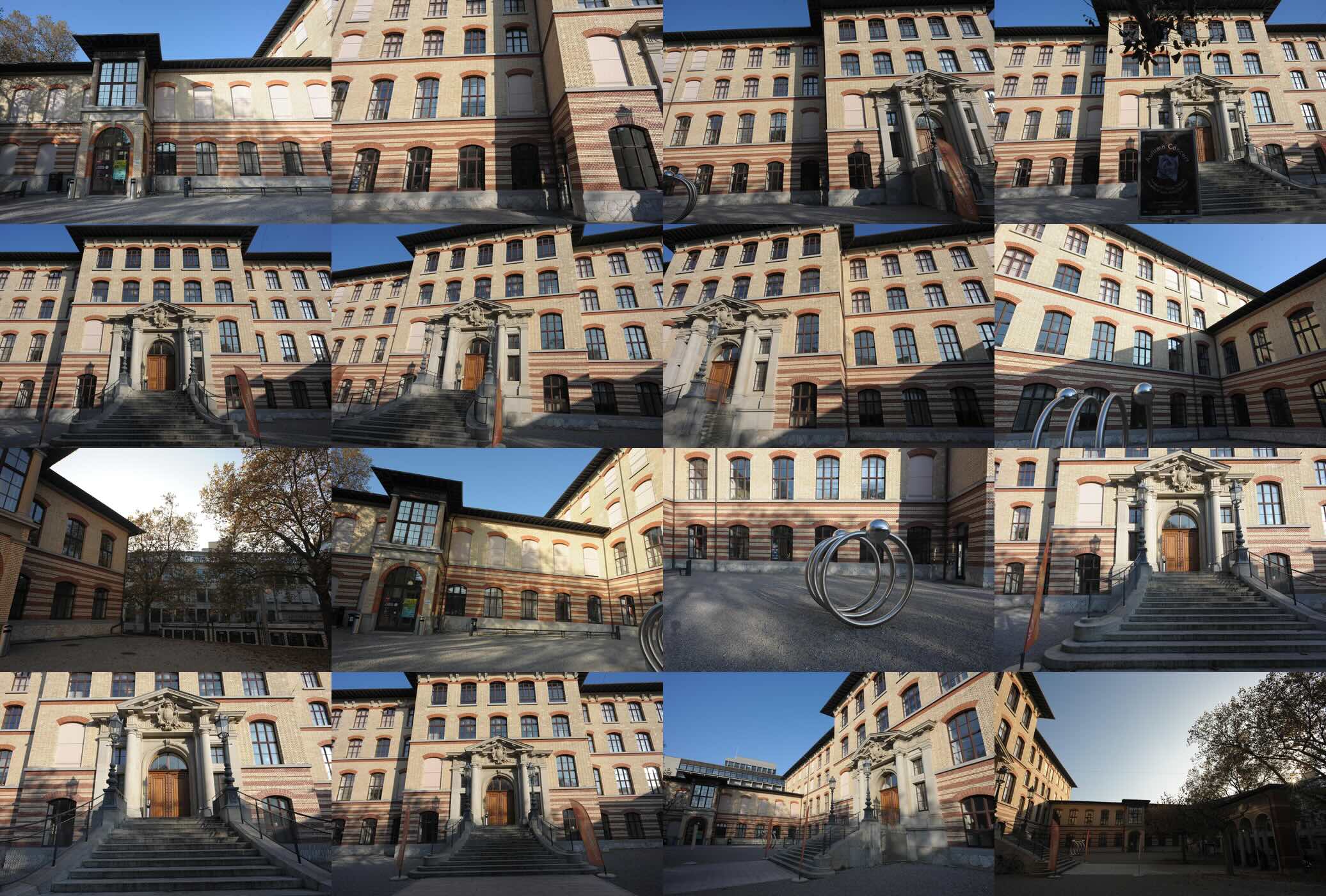}
    \end{subfigure}\hfill
    \begin{subfigure}{0.16\textwidth}
        \includegraphics[width=\linewidth]{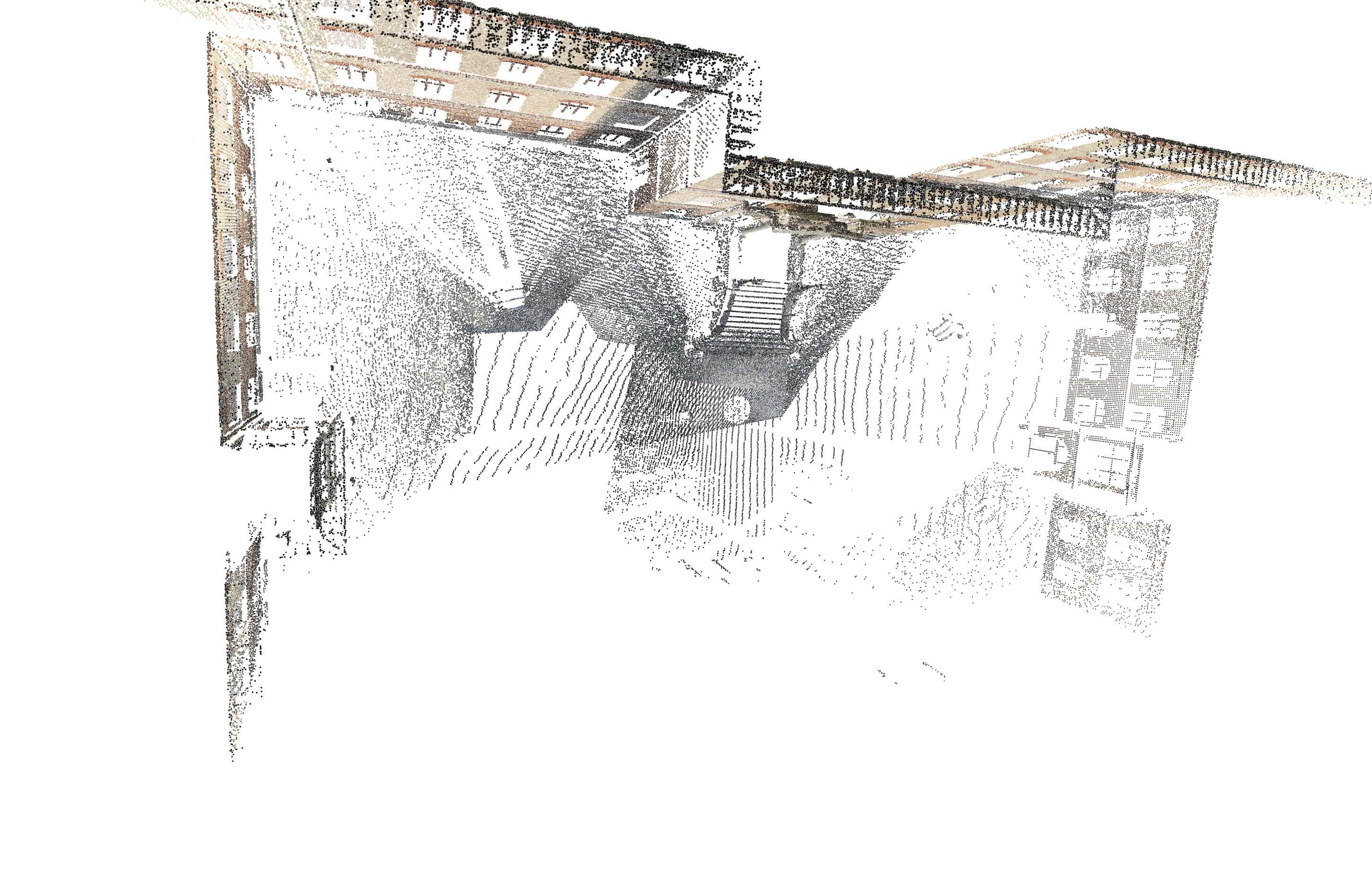}
    \end{subfigure}\hfill
    \begin{subfigure}{0.16\textwidth}
        \includegraphics[width=\linewidth]{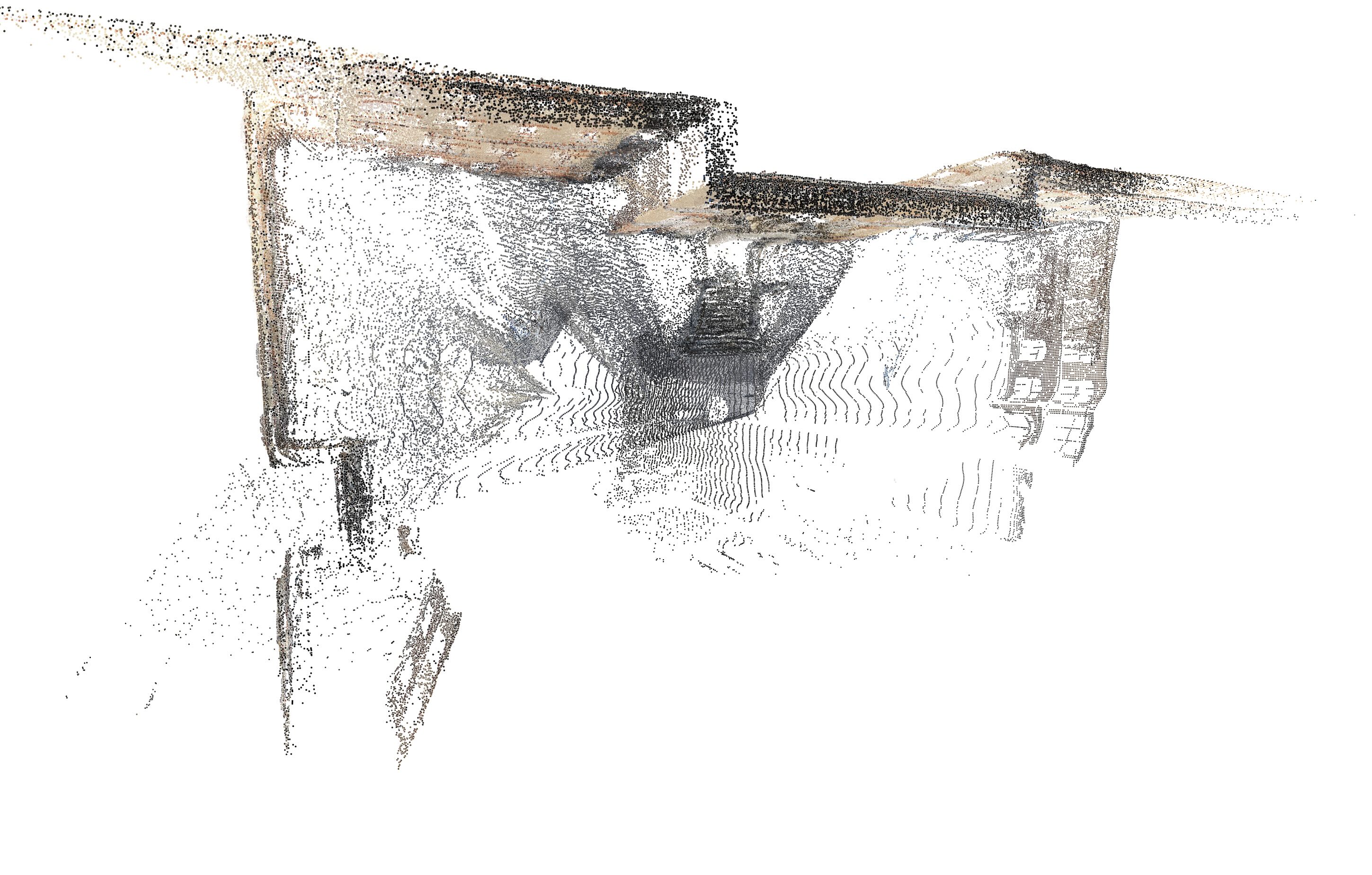}
    \end{subfigure}\hfill
    \begin{subfigure}{0.16\textwidth}
        \includegraphics[width=\linewidth]{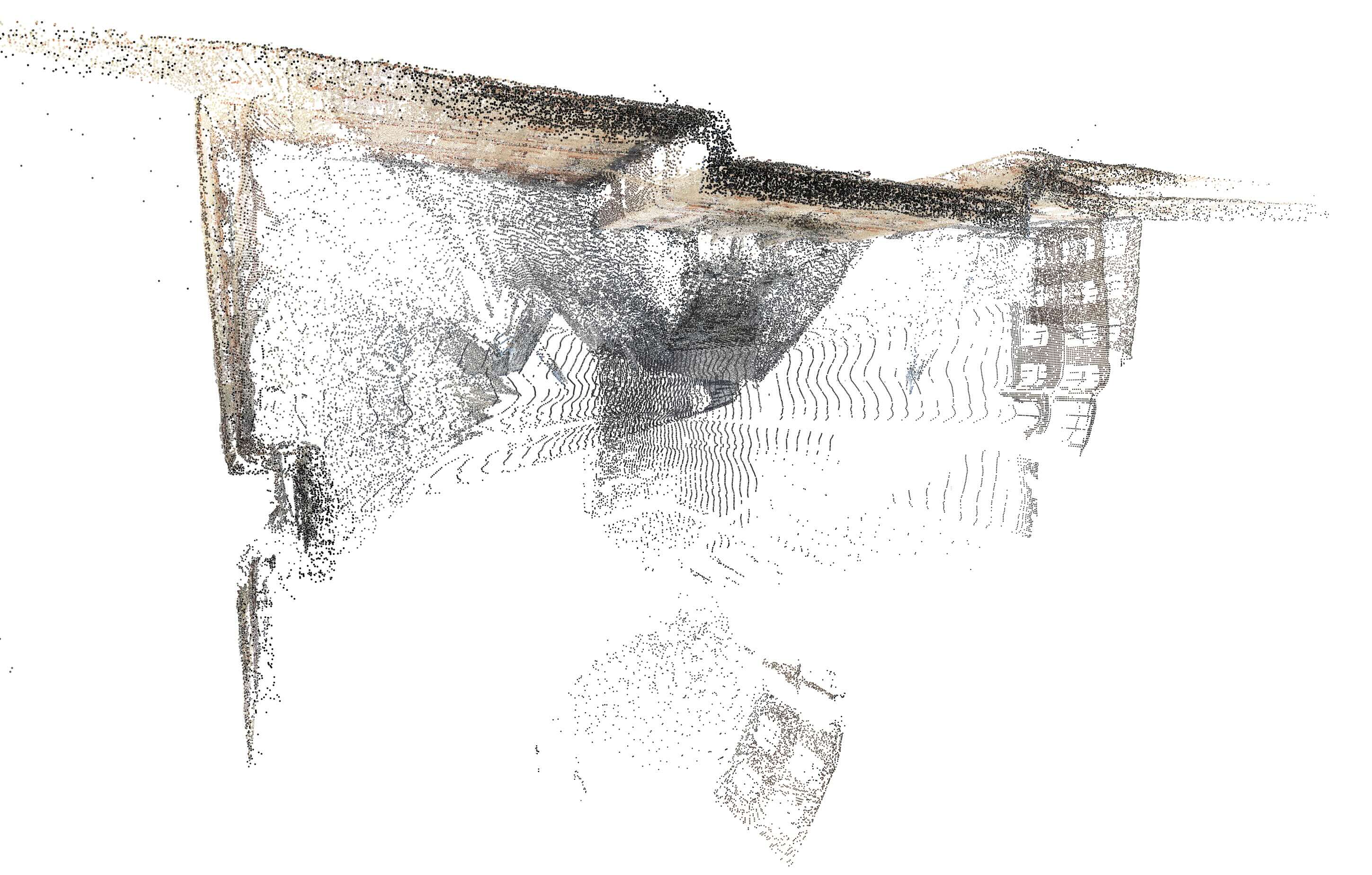}
    \end{subfigure}\hfill
    \begin{subfigure}{0.16\textwidth}
        \includegraphics[width=\linewidth]{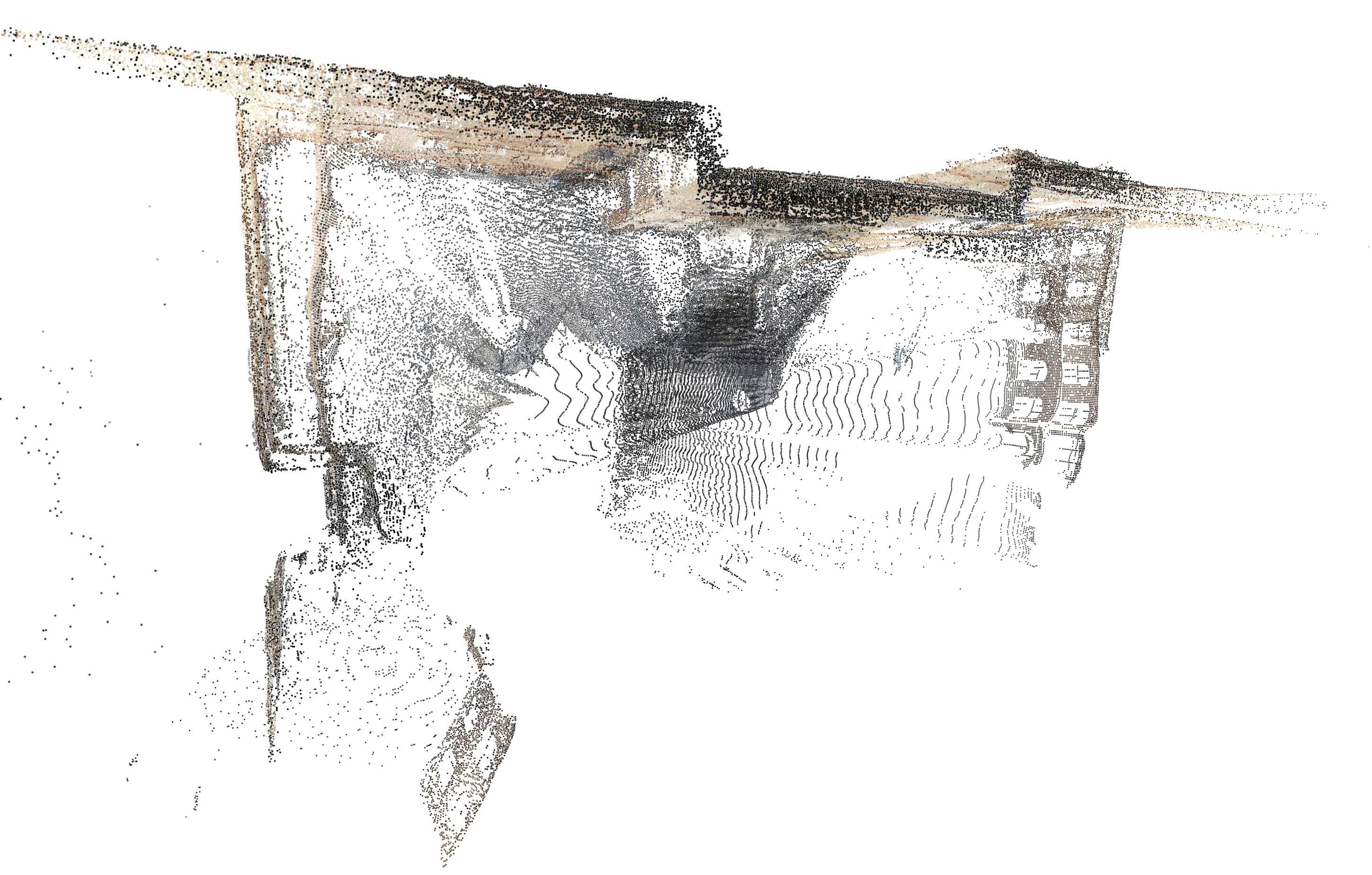}
    \end{subfigure}\hfill
    \begin{subfigure}{0.16\textwidth}
        \includegraphics[width=\linewidth]{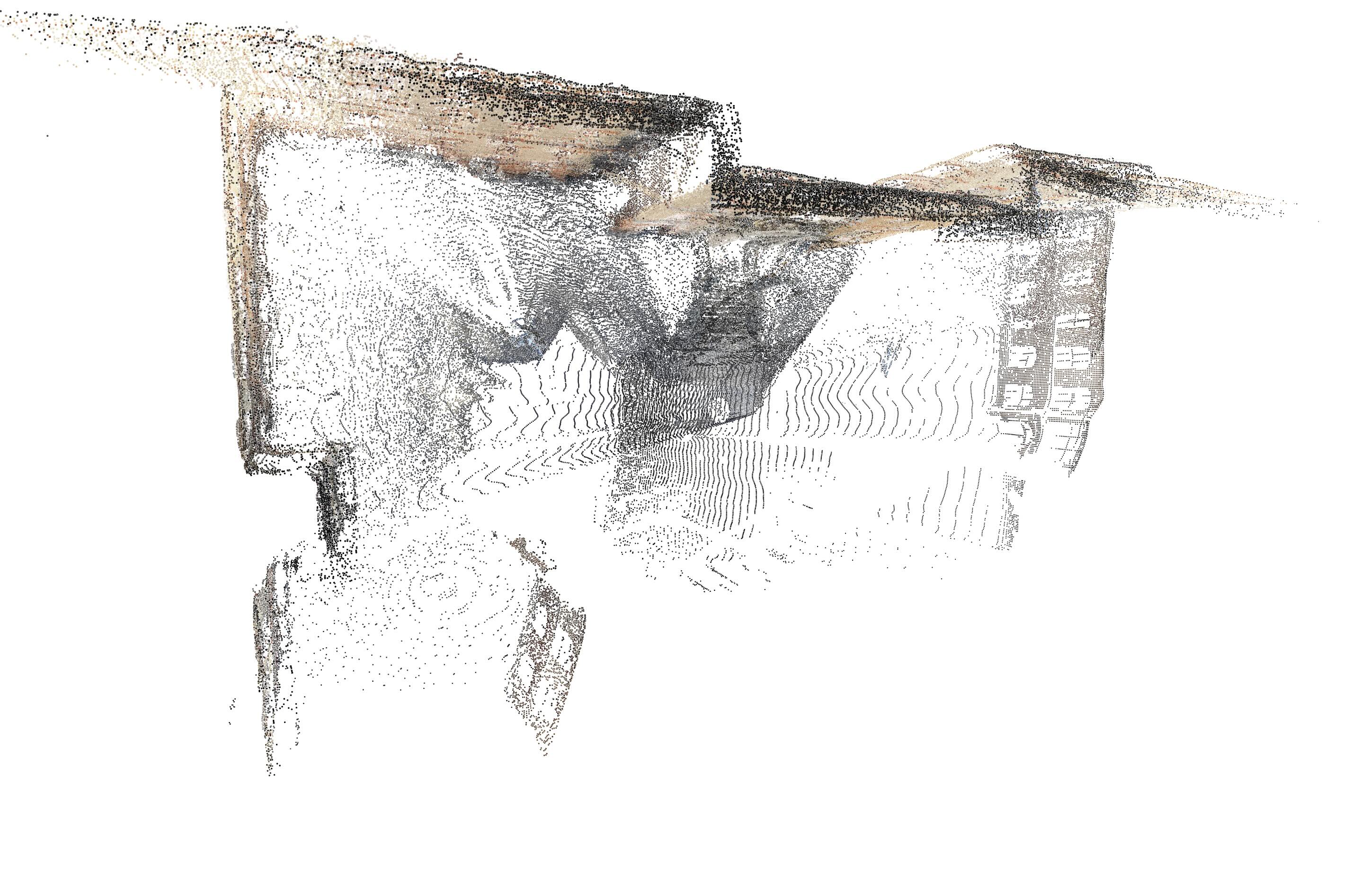}
    \end{subfigure}

    \vspace{1em} 

    \rotatebox{90}{\makebox[4em][c]{$\Pi^3$}}%
    \hspace{0.5em}
    \begin{subfigure}{0.16\textwidth}
        \includegraphics[width=\linewidth]{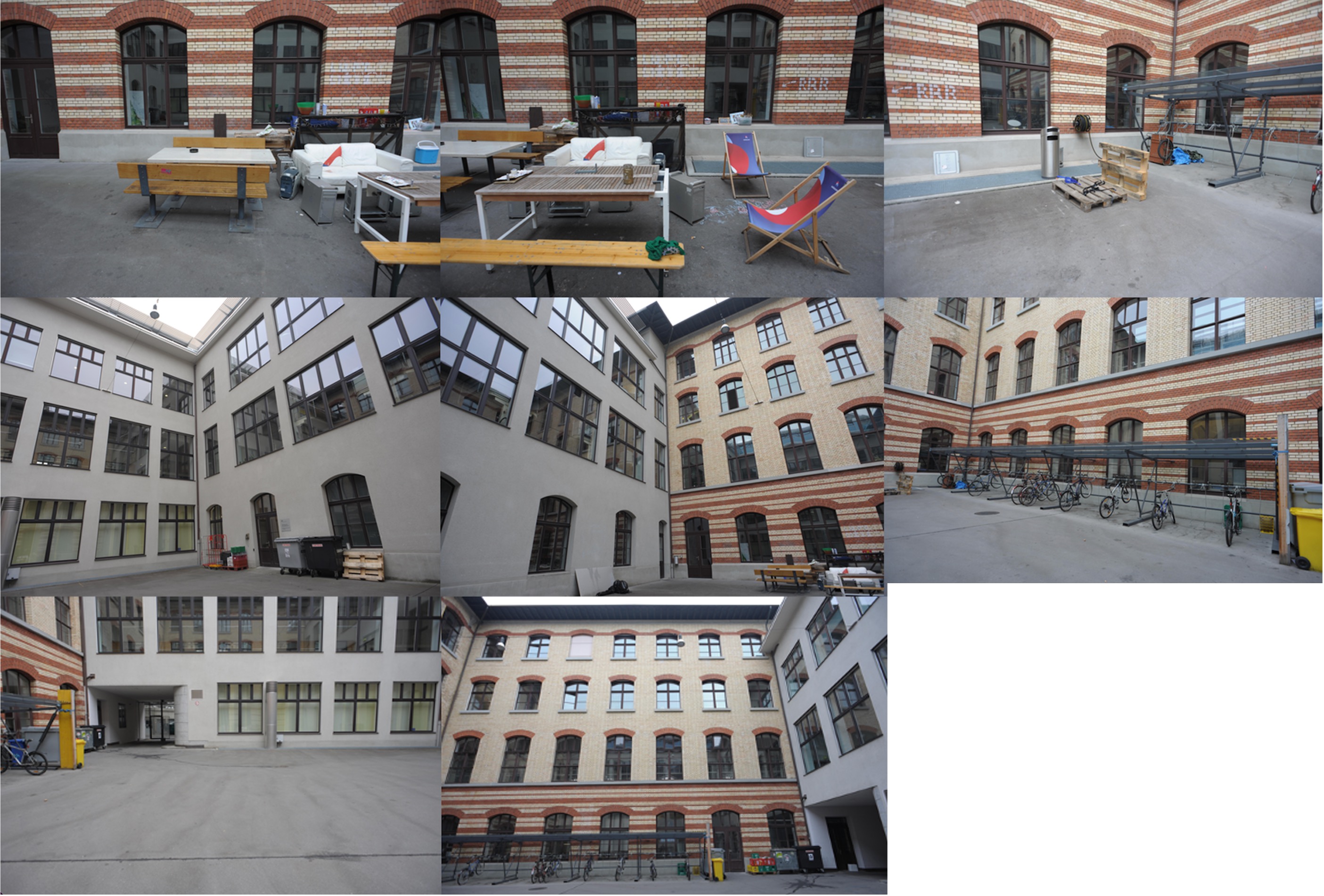}
    \end{subfigure}\hfill
    \begin{subfigure}{0.16\textwidth}
        \includegraphics[width=\linewidth]{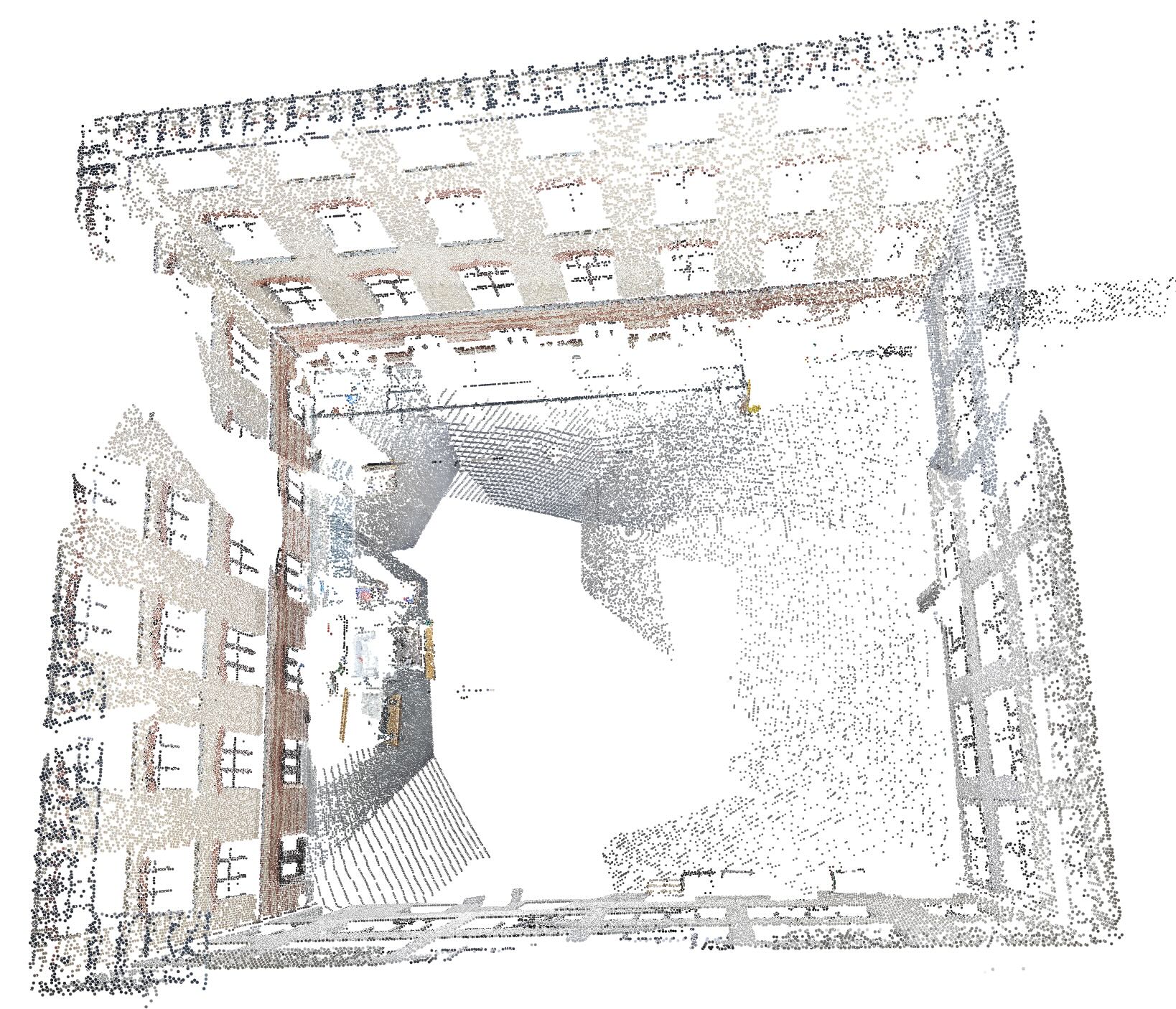}
    \end{subfigure}\hfill
    \begin{subfigure}{0.16\textwidth}
        \includegraphics[width=\linewidth]{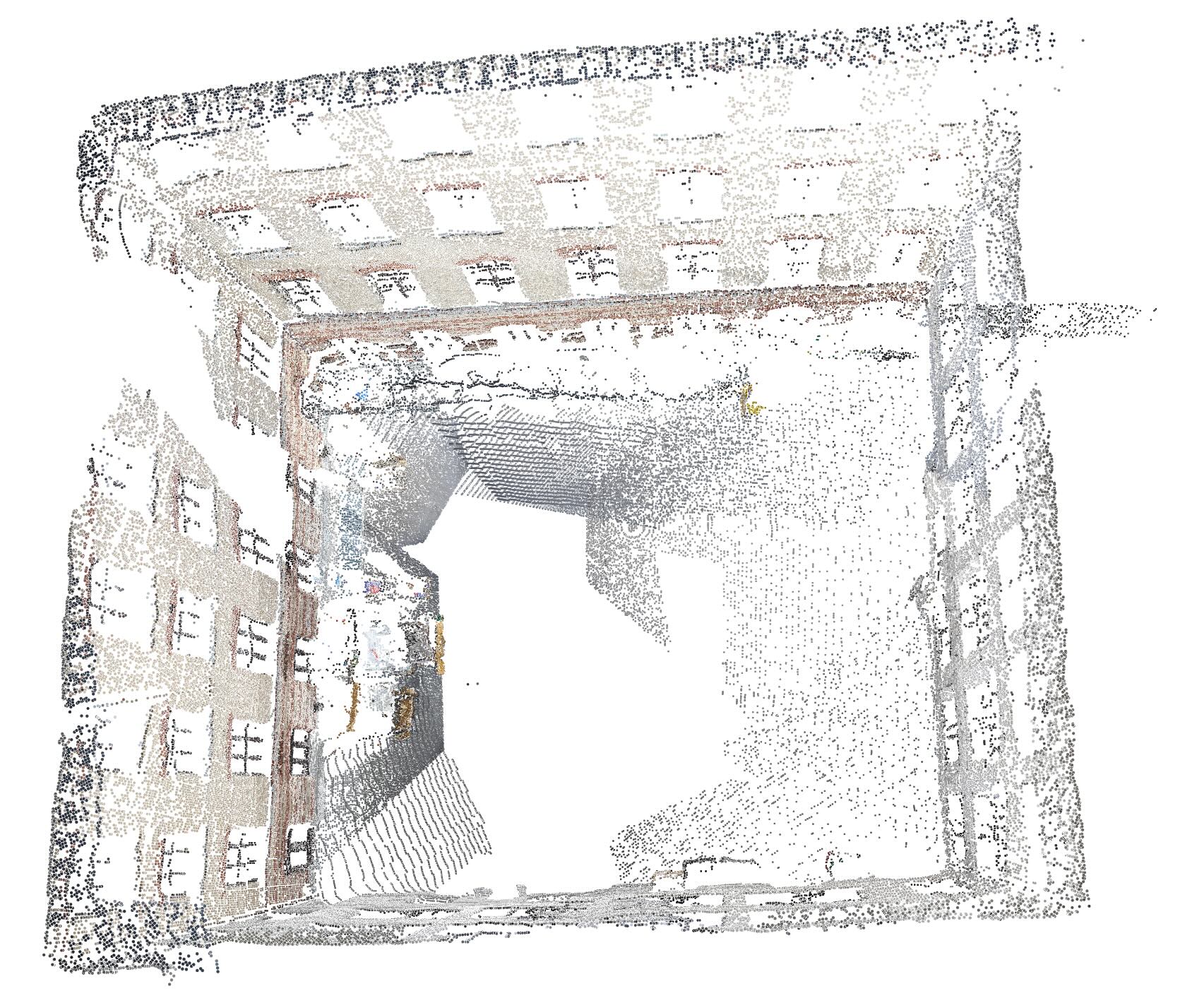}
    \end{subfigure}\hfill
    \begin{subfigure}{0.16\textwidth}
        \includegraphics[width=\linewidth]{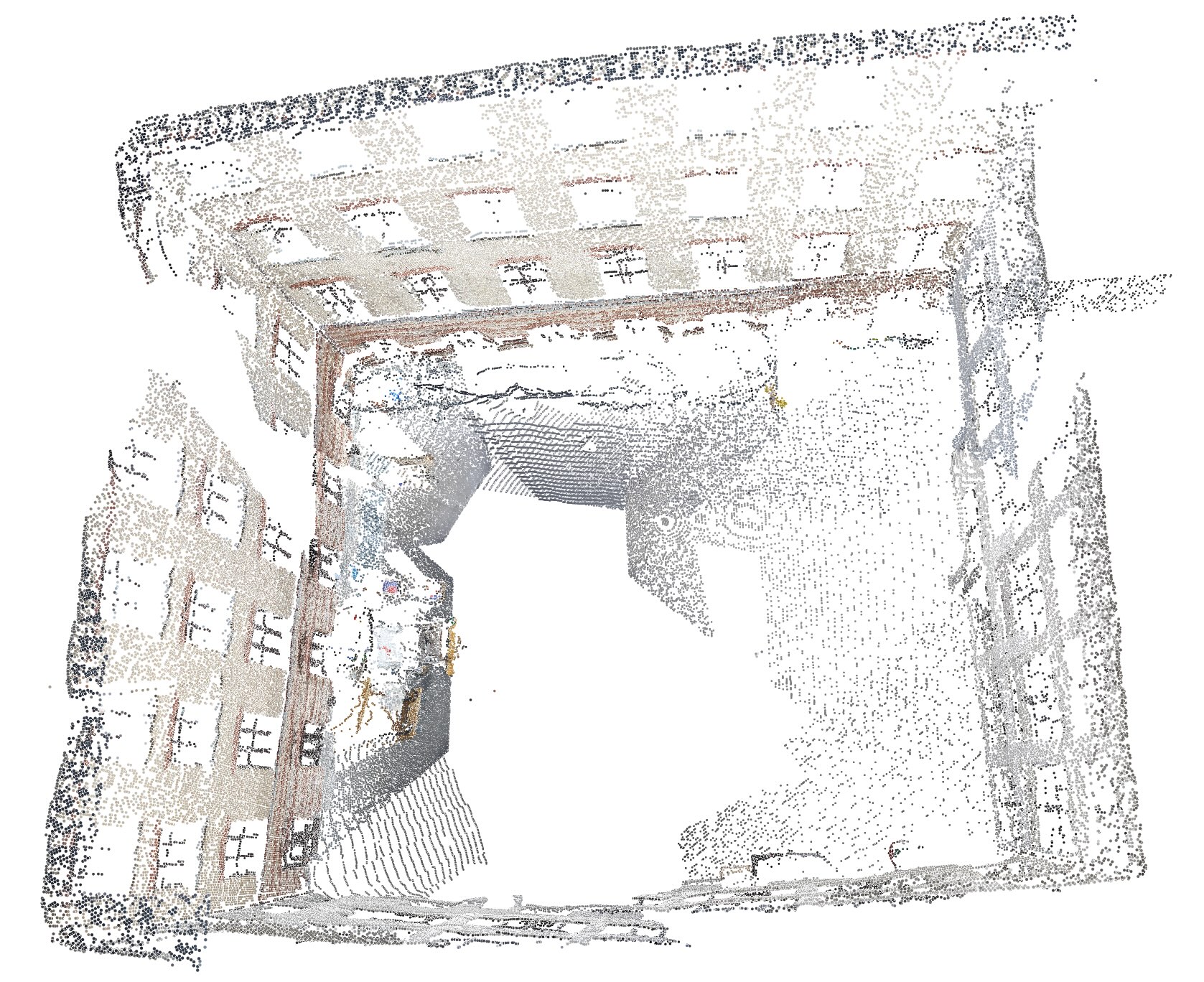}
    \end{subfigure}\hfill
    \begin{subfigure}{0.16\textwidth}
        \includegraphics[width=\linewidth]{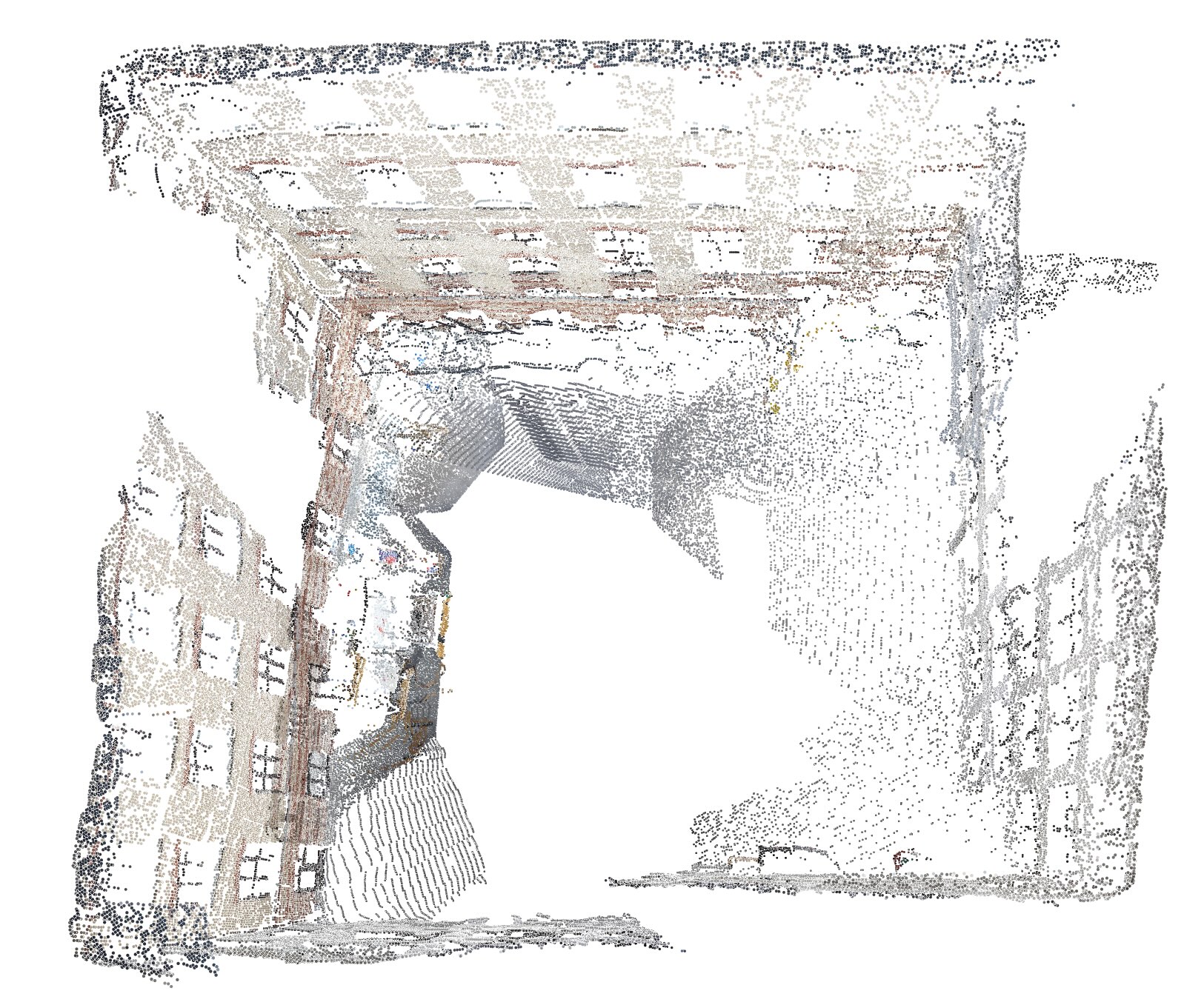}
    \end{subfigure}\hfill
    \begin{subfigure}{0.16\textwidth}
        \includegraphics[width=\linewidth]{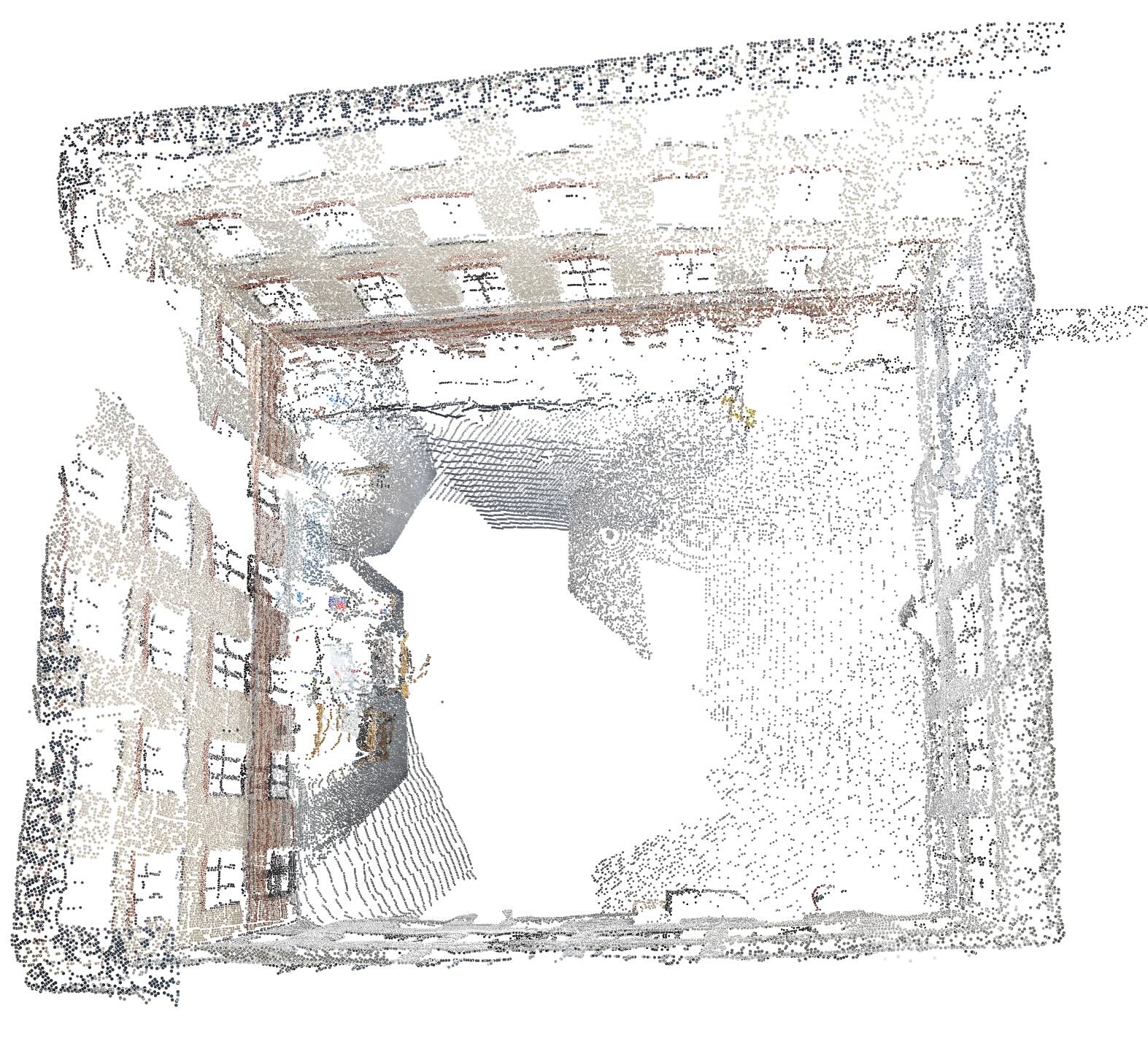}
    \end{subfigure}

    \caption{\textbf{Qualitative comparison of predicted point clouds from our method and baselines.} Compared with sparse baselines, our method produces visually more accurate reconstructions and reduces the multi-layer wall phenomenon across both architectures.}
    \label{fig:full_comparison}
\end{figure*}

Following $\pi^3$~\cite{wang2025pi3}, we compare our predicted point cloud with baseline methods. We report the mean and median of accuracy, completeness, and normal consistency. As shown in Table~\ref{tab:comparison_results} and illustrated in Figure~\ref{fig:full_comparison}, our proposed Speed3R method demonstrates a superior trade-off between performance and efficiency for pointmap estimation. When compared against other sparse methods, Speed3R consistently achieves the best results across nearly all metrics on both the DTU and ETH3D datasets, establishing it as the state-of-the-art among efficiency-focused techniques. More importantly, while the dense baseline models (VGGT and $\pi^3$) exhibit slightly better accuracy, our Speed3R variants remain highly competitive, incurring only a marginal performance degradation. This indicates that the learned sparse attention patterns effectively preserve the most critical information for high-quality reconstruction.

\subsection{Ablation Study}

\begin{table}[!tp]
\centering
\caption{\textbf{Ablation study for Speed3R-$\boldsymbol{\pi^3}$.} The ``Time" column reports the average inference time on the T\&T~\cite{Knapitsch2017} dataset.}

\label{tab:ablation_pi3}
\adjustbox{valign=t,width=\linewidth}{
\sisetup{detect-all=true,detect-weight=true}
\begin{tabular}{lccc}
\toprule
Method & \makecell{RE10K~\cite{zhou2018stereo} \\ AUC@30 $\uparrow$} & \makecell{T\&T~\cite{Knapitsch2017} \\ AUC@30 $\uparrow$} & \makecell{Time \\ s$\downarrow$} \\
\midrule

Base & 86.35 & 78.69 & 4.19 \\
(1) w/o Compress. Branch Value & 86.29 & 77.90 & 3.99 \\
(2) w/o Select. Branch & 83.44 & 76.84 & 3.56 \\
(3) w/ register & 86.39 & 78.57 & 4.25  \\
\hline

(4) Top-8 & 85.37 & 78.17 & 3.72 \\
(5) Top-16 & 85.98 & 78.55 & 3.92 \\
(6) Top-64 & 86.42 & 78.90 & 4.64 \\
\hline
(7) 8x8 window & 86.49 & 78.71 & 5.27 \\
\hline
(8) w/o distillation & 85.18 & 77.81 &  4.19 \\
\bottomrule
\end{tabular}
}
\end{table}
\begin{table}[!tp]
\centering
\caption{\textbf{Ablation Study of Speed3R-VGGT.} Analysis of the selection attention branch of GSA in Speed3R-VGGT.}

\label{tab:ablation_vggt}
\adjustbox{valign=t,width=\linewidth}{
\sisetup{detect-all=true,detect-weight=true}
\begin{tabular}{lccc}
\toprule
Method & \makecell{RE10K~\cite{zhou2018stereo} \\ AUC@30 $\uparrow$} & \makecell{T\&T~\cite{Knapitsch2017} \\ AUC@30 $\uparrow$} & \makecell{Time \\ s$\downarrow$} \\
\midrule

Base & 74.56 & 76.57 & 6.52 \\
(1) w/o reference frame attn. & 73.87 & 75.69 & 5.91 \\
(2) w/o register token & 74.14 & 76.21 & 6.50 \\
(3) w/ all special token & 74.91 & 74.77 & 7.05 \\
\bottomrule
\end{tabular}
}
\end{table}

We first conduct ablation studies on the Speed3R-$\pi^3$, with a baseline using a 4x4 window and top-32. We train all models with 40 epochs and gradient accumulation factor of 2. As detailed in Table~\ref{tab:ablation_pi3}, we first analyze our GSA module. Removing the compression branch value (1) impairs long-sequence performance on the T\&T~\cite{Knapitsch2017} dataset by sacrificing global context, while removing the selection branch (2) is uniformly detrimental, causing a significant drop on both datasets. Adding a register token (3) has a negligible effect. In our hyperparameter analysis(4-7), we find that our choice (4x4 window with top-32 indices) strikes the balance between accuracy and efficiency. Finally, removing knowledge distillation (8) substantially degrades performance across both datasets. This result highlights its importance in mitigating the impact of noisy labels from the real-world dataset.

Due to the utilization of reference-frame and camera token of VGGT~\cite{wang2025vggt}, the design of Speed3R-VGGT differs from that of Speed3R-$\pi^3$, we ablate the special design choice of the \textit{selection branch} of Speed3R-VGGT in Table~\ref{tab:ablation_vggt}. Removing the reference frame attention (1) Impairs performance on both datasets, reflecting the inherent inductive bias of the original VGGT. Following Speed3R-$\pi^3$,  we also try to remove the register token. However we find that removing the register tokens (2) also degrades performance. We find these tokens provide crucial stability by acting as an assistant to the camera token; while it is not used for the final pose prediction, its absence clearly impairs model accuracy. Finally, forcing patch tokens to attend to all special tokens hurts long-sequence performance. We hypothesize that for longer sequences, these appended special tokens will dominate over the more crucial top-k patch selection, thereby impairing performance.

\subsection{Latency Benchmarking}
\begin{figure}[h]
\centering
    \caption{\textbf{Inference Time for Different Models.} Mean latency (in seconds) for varying sequence lengths. Our method achieves a 12.4× speedup on sequences of 1024 images.}
    \label{fig:latency}

\resizebox{\linewidth}{!}{%
\begin{tabular}{lcccccc}
\toprule
\textbf{Seq Length} & \textbf{32} & \textbf{64} & \textbf{128} & \textbf{256} & \textbf{512} & \textbf{1024} \\
\midrule
Full Attn.($\pi^3$) & 0.50 & 1.31 & 3.97 & 13.41 & 50.01 & 202.39 \\
Block Sparse~\cite{wang2025faster} & 0.46 & 0.85 & 1.69 & 3.77 & 9.64 & 29.58 \\
FastVGGT~\cite{shen2025fastvggt} & 0.44 & 0.88 & 1.96 & 4.95 & 14.13 & 45.49 \\
Ours & 0.37 & 0.71 & 1.44 & 3.06 & 6.83 & 16.38 \\
\bottomrule
\end{tabular}%
}
    \vspace{1em}
    \includegraphics[width=1.0\linewidth]{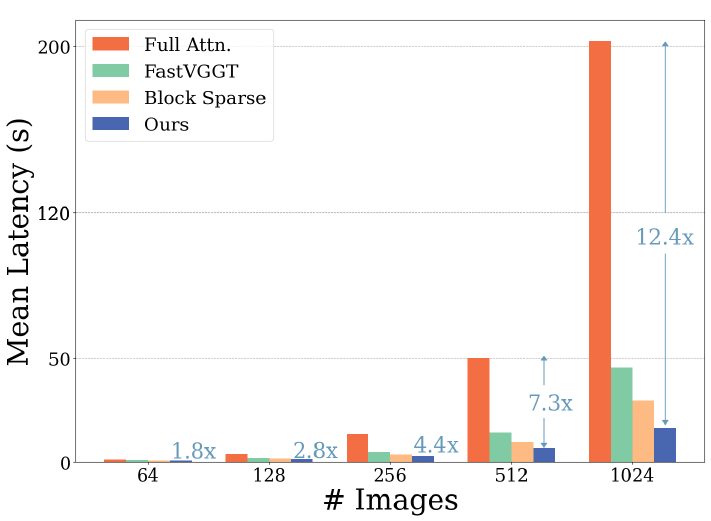}
\end{figure}

As shown in Figure~\ref{fig:latency}, the Speed3R-$\pi^3$ model demonstrates superior inference speed compared to all baselines. This efficiency arises from its computational complexity, which avoids the quadratic $O(n^2)$ of the standard full-attention baseline. The performance advantage becomes increasingly pronounced with sequence length, reaching a $12.4 \times$ speedup at an input length of 1024. This result highlights our model's efficacy for high-throughput, long-sequence processing. The benchmarking results for Speed3R-VGGT can be found in the supplementary materials.
\section{Discussions}
\label{sec:discussion}

\paragraph{Dense \textit{vs} Sparse Attention.} While our method establishes a new efficiency-accuracy \textit{Pareto} frontier, the accuracy of our sparse model on short sequences does not yet match its dense counterparts. We attribute this gap primarily to limitations in data and computational resources. To investigate this, we train both the full-attention and our sparse-attention models under identical conditions (Details in Supplementary). As illustrated in Figure~\ref{fig:loss}, our sparse model achieves a comparable training loss to the full-attention model, while completing the training process 1.12x faster. This suggests that the models possess similar learning capacities. Therefore, despite not yet exceeding the dense model's accuracy, Speed3R demonstrates a viable path toward greater efficiency.

\paragraph{Test-time adaptation.}
Although our flagship model is trained with top-32, we observe that increasing the top-k value to top-64 and top-128 during inference consistently improves performance on long-sequence datasets (T\&T~\cite{Knapitsch2017}) in Table~\ref{tab:tttt}. Notably, this adjustment enables our model to outperform dense models on RTA@5 and AUC@30 during testing, highlighting the robustness of our method and its flexibility in handling long sequences.

\paragraph{Sparsification Challenges.}
The nature of reconstruction tasks distinguishes them from domains where sparse methods have flourished, such as Large Language Models (LLMs)\cite{yuan2025native, lu2025moba} and generation models\cite{zhang2025vsa, wu2025direct3ds2}. The pose regression requires high numerical precision and is extremely sensitive. This contrasts with the probabilistic and often perceptual objectives of text or image generation. While our sparse method achieves comparable accuracy on the AUC@30 pose estimation metric, it still underperforms relative to the dense model at the stricter AUC@5 threshold. Recognizing these domain-specific challenges, our work presents an initial exploration into developing a sparse attention mechanism, providing a foundational step towards reconciling efficiency with precision. 


\paragraph{Limitations.} The dual-branch architecture of GSA, while enabling sparsity, incurs a 15\% memory overhead compared to full attention. In practice, this is manageable, as the model can accommodate up to 1024 images on an 80GB GPU. Looking forward, the strategy proposed in SAIL-Recon~\cite{deng2025sail} provides a path to extend our method to arbitrarily long sequences, removing the memory constraint.
\begin{table}[!tp]
\begin{center}
\caption{\textbf{Test-time adaptation on Tanks \& Temples.~\citep{Knapitsch2017}.} }
\label{tab:tttt}
\adjustbox{valign=t,width=\linewidth}{
\begin{tabular}{lcccc}
\toprule
 Method  & RRA@5 $\uparrow$ & RTA@5 $\uparrow$ & AUC@30 $\uparrow$ & Time [s] $\downarrow$ \\
\midrule
$\pi^3$~\cite{wang2025pi3} & \textcolor{gray}{72.14} & \textcolor{gray}{81.26} & \textcolor{gray}{79.63} & \textcolor{gray}{22.32} \\
Speed3R-$\pi^3$(top-8)    & 69.73 & 77.60 & 78.21 & 3.72 \\
Speed3R-$\pi^3$(top-16)   & 70.26 & 79.49 & 79.21 & 3.92 \\
Speed3R-$\pi^3$(top-32)   & 70.72 & 80.72 & 79.77 & 4.19 \\
Speed3R-$\pi^3$(top-64)   & 71.60 & 81.54 & 80.10 & 4.64 \\
Speed3R-$\pi^3$(top-128)  & 71.89 & 82.00 & 80.33 & 6.07 \\

\bottomrule
\end{tabular}
}
\end{center}
\end{table}
\begin{figure}[t]
\centering
    \caption{\textbf{Training loss curve of dense attention and our GSA}.}
    \includegraphics[width=1.0\linewidth]{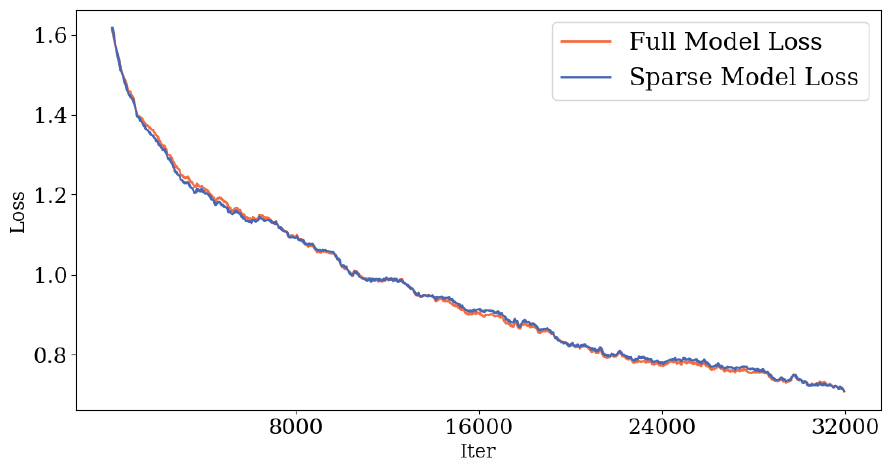}
    \label{fig:loss}
\end{figure}

\section{Conclusion}
\label{sec:conclusion}

In this paper, we introduced Speed3R, a novel sparse attention model designed to mitigate the prohibitive computational cost of feed-forward 3D reconstruction. Inspired by the efficiency of classical SfM and recent advancements in sparse attention, Speed3R employs a trainable dual-branch mechanism to focus computation on a small subset of informative tokens. Our approach establishes a new SoTA in the efficiency-accuracy trade-off, achieving a \textbf{12.4x speedup} on 1024-images sequences with minimal impact on geometric accuracy. We validated the robustness of Speed3R by integrating it with two backbones, where it consistently outperformed training-free alternatives, paving the way for practical and scalable large-scale 3D scene modeling.

\section{Acknowledgement}
This work is supported by Hong Kong Research Grant Council - General Research Fund (Grant No. 17213825) and HKU Seed Fund for PI Research. The authors would like to thank Hongjun Wang for insightful discussions regarding kernel implementation, and Zhiheng Wu and Yumeng Zhang for generously sharing GPU resources during the computationally intensive phases of the experiments.
{
    \small
    \bibliographystyle{ieeenat_fullname}
    \bibliography{main,refs}

\begin{thebibliography}{56}
\providecommand{\natexlab}[1]{#1}
\providecommand{\url}[1]{\texttt{#1}}
\expandafter\ifx\csname urlstyle\endcsname\relax
  \providecommand{\doi}[1]{doi: #1}\else
  \providecommand{\doi}{doi: \begingroup \urlstyle{rm}\Url}\fi

\bibitem[Baruch et~al.(2021)Baruch, Chen, Dehghan, Dimry, Feigin, Fu, Gebauer, Joffe, Kurz, Schwartz, and Shulman]{dehghan2021arkitscenes}
Gilad Baruch, Zhuoyuan Chen, Afshin Dehghan, Tal Dimry, Yuri Feigin, Peter Fu, Thomas Gebauer, Brandon Joffe, Daniel Kurz, Arik Schwartz, and Elad Shulman.
\newblock {ARK}itscenes - a diverse real-world dataset for 3d indoor scene understanding using mobile {RGB}-d data.
\newblock In \emph{NeurIPS}, 2021.

\bibitem[Bay et~al.(2008)Bay, Ess, Tuytelaars, and Van~Gool]{bay2008speeded}
Herbert Bay, Andreas Ess, Tinne Tuytelaars, and Luc Van~Gool.
\newblock Speeded-up robust features (surf).
\newblock \emph{Computer vision and image understanding}, 2008.

\bibitem[Beltagy et~al.(2020)Beltagy, Peters, and Cohan]{Beltagy2020Longformer}
Iz Beltagy, Matthew~E. Peters, and Arman Cohan.
\newblock Longformer: The long-document transformer.
\newblock \emph{arXiv preprint arXiv:2004.05150}, 2020.

\bibitem[Cabon et~al.(2020{\natexlab{a}})Cabon, Murray, and Humenberger]{cabon2020virtual}
Yohann Cabon, Naila Murray, and Martin Humenberger.
\newblock Virtual kitti 2.
\newblock \emph{arXiv preprint arXiv:2001.10773}, 2020{\natexlab{a}}.

\bibitem[Cabon et~al.(2020{\natexlab{b}})Cabon, Murray, and Humenberger]{cabon2020vkitti2}
Yohann Cabon, Naila Murray, and Martin Humenberger.
\newblock Virtual kitti 2, 2020{\natexlab{b}}.

\bibitem[Cai et~al.(2025)Cai, Yang, Zhang, Guo, Xiao, Yang, Xu, Yang, Yuille, Guibas, Agrawala, Jiang, and Wetzstein]{cai2025moc}
Shengqu Cai, Ceyuan Yang, Lvmin Zhang, Yuwei Guo, Junfei Xiao, Ziyan Yang, Yinghao Xu, Zhenheng Yang, Alan Yuille, Leonidas Guibas, Maneesh Agrawala, Lu Jiang, and Gordon Wetzstein.
\newblock Mixture of contexts for long video generation.
\newblock 2025.

\bibitem[Dai et~al.(2017)Dai, Chang, Savva, Halber, Funkhouser, and Nie{\ss}ner]{dai2017scannet}
Angela Dai, Angel~X. Chang, Manolis Savva, Maciej Halber, Thomas Funkhouser, and Matthias Nie{\ss}ner.
\newblock Scannet: Richly-annotated 3d reconstructions of indoor scenes.
\newblock In \emph{CVPR}, 2017.

\bibitem[Dao(2024)]{dao2023flashattention2}
Tri Dao.
\newblock Flash{A}ttention-2: Faster attention with better parallelism and work partitioning.
\newblock In \emph{ICLR}, 2024.

\bibitem[Deng et~al.(2025)Deng, Li, Xie, Ren, Zhang, Tan, and Guo]{deng2025sail}
Junyuan Deng, Heng Li, Tao Xie, Weiqiang Ren, Qian Zhang, Ping Tan, and Xiaoyang Guo.
\newblock Sail-recon: Large sfm by augmenting scene regression with localization.
\newblock \emph{arXiv preprint arXiv:2508.17972}, 2025.

\bibitem[DeTone et~al.(2018)DeTone, Malisiewicz, and Rabinovich]{detone2018superpoint}
Daniel DeTone, Tomasz Malisiewicz, and Andrew Rabinovich.
\newblock Superpoint: Self-supervised interest point detection and description.
\newblock In \emph{CVPR}, 2018.

\bibitem[Dosovitskiy et~al.(2021)Dosovitskiy, Beyer, Kolesnikov, Weissenborn, Zhai, Unterthiner, Dehghani, Minderer, Heigold, Gelly, Uszkoreit, and Houlsby]{dosovitskiy2020vit}
Alexey Dosovitskiy, Lucas Beyer, Alexander Kolesnikov, Dirk Weissenborn, Xiaohua Zhai, Thomas Unterthiner, Mostafa Dehghani, Matthias Minderer, Georg Heigold, Sylvain Gelly, Jakob Uszkoreit, and Neil Houlsby.
\newblock An image is worth 16x16 words: Transformers for image recognition at scale.
\newblock \emph{ICLR}, 2021.

\bibitem[Gao et~al.(2024)Gao, Zeng, Du, Cao, Zhou, Qi, Lai, So, Cao, Yang, et~al.]{gao2024seerattention}
Yizhao Gao, Zhichen Zeng, Dayou Du, Shijie Cao, Peiyuan Zhou, Jiaxing Qi, Junjie Lai, Hayden Kwok-Hay So, Ting Cao, Fan Yang, et~al.
\newblock Seerattention: Learning intrinsic sparse attention in your llms.
\newblock \emph{arXiv preprint arXiv:2410.13276}, 2024.

\bibitem[Hartley and Zisserman(2000)]{hartley_multiple_2000}
Richard Hartley and Andrew Zisserman.
\newblock \emph{Multiple {View} {Geometry} in {Computer} {Vision}}.
\newblock Cambridge University Press, 2000.

\bibitem[He et~al.(2015)He, Zhang, Ren, and Sun]{he2015delving}
Kaiming He, Xiangyu Zhang, Shaoqing Ren, and Jian Sun.
\newblock Delving deep into rectifiers: Surpassing human-level performance on imagenet classification.
\newblock In \emph{ICCV}, 2015.

\bibitem[He et~al.(2025)He, Zou, Chen, Guo, Liang, Yuan, Ouyang, Cao, and Li]{he2025triposf}
Xianglong He, Zi-Xin Zou, Chia-Hao Chen, Yuan-Chen Guo, Ding Liang, Chun Yuan, Wanli Ouyang, Yan-Pei Cao, and Yangguang Li.
\newblock Sparseflex: High-resolution and arbitrary-topology 3d shape modeling.
\newblock \emph{arXiv preprint arXiv:2503.21732}, 2025.

\bibitem[Jensen et~al.(2014)Jensen, Dahl, Vogiatzis, Tola, and Aan{\ae}s]{dtudataset}
Rasmus Jensen, Anders Dahl, George Vogiatzis, Engil Tola, and Henrik Aan{\ae}s.
\newblock Large scale multi-view stereopsis evaluation.
\newblock In \emph{CVPR}, 2014.

\bibitem[Knapitsch et~al.(2017)Knapitsch, Park, Zhou, and Koltun]{Knapitsch2017}
Arno Knapitsch, Jaesik Park, Qian-Yi Zhou, and Vladlen Koltun.
\newblock Tanks and temples: Benchmarking large-scale scene reconstruction.
\newblock \emph{ACM TOG}, 2017.

\bibitem[Leroy et~al.(2024)Leroy, Cabon, and Revaud]{mast3r_eccv24}
Vincent Leroy, Yohann Cabon, and Jerome Revaud.
\newblock Grounding image matching in 3d with mast3r, 2024.

\bibitem[Lindenberger et~al.(2023)Lindenberger, Sarlin, and Pollefeys]{lindenberger2023lightglue}
Philipp Lindenberger, Paul-Edouard Sarlin, and Marc Pollefeys.
\newblock Lightglue: Local feature matching at light speed.
\newblock \emph{arXiv preprint arXiv:2306.13643}, 2023.

\bibitem[Ling et~al.(2024)Ling, Sheng, Tu, Zhao, Xin, Wan, Yu, Guo, Yu, Lu, et~al.]{ling2024dl3dv}
Lu Ling, Yichen Sheng, Zhi Tu, Wentian Zhao, Cheng Xin, Kun Wan, Lantao Yu, Qianyu Guo, Zixun Yu, Yawen Lu, et~al.
\newblock Dl3dv-10k: A large-scale scene dataset for deep learning-based 3d vision.
\newblock In \emph{CVPR}, pages 22160--22169, 2024.

\bibitem[Liu et~al.(2025)Liu, Li, Zhao, Zhang, and Guo]{liu2025clusterkv}
Guangda Liu, Chengwei Li, Jieru Zhao, Chenqi Zhang, and Minyi Guo.
\newblock Clusterkv: Manipulating llm kv cache in semantic space for recallable compression.
\newblock In \emph{2025 62nd ACM/IEEE Design Automation Conference (DAC)}, 2025.

\bibitem[Lowe(1999)]{lowe1999object}
David~G Lowe.
\newblock Object recognition from local scale-invariant features.
\newblock In \emph{ICCV}, 1999.

\bibitem[Lu et~al.(2025)Lu, Jiang, Liu, Du, Jiang, Hong, Liu, He, Yuan, Wang, et~al.]{lu2025moba}
Enzhe Lu, Zhejun Jiang, Jingyuan Liu, Yulun Du, Tao Jiang, Chao Hong, Shaowei Liu, Weiran He, Enming Yuan, Yuzhi Wang, et~al.
\newblock Moba: Mixture of block attention for long-context llms.
\newblock \emph{arXiv preprint arXiv:2502.13189}, 2025.

\bibitem[Oquab et~al.(2023)Oquab, Darcet, Moutakanni, Vo, Szafraniec, Khalidov, Fernandez, Haziza, Massa, El-Nouby, et~al.]{oquab2023dinov2}
Maxime Oquab, Timoth{\'e}e Darcet, Th{\'e}o Moutakanni, Huy Vo, Marc Szafraniec, Vasil Khalidov, Pierre Fernandez, Daniel Haziza, Francisco Massa, Alaaeldin El-Nouby, et~al.
\newblock Dinov2: Learning robust visual features without supervision.
\newblock \emph{arXiv preprint arXiv:2304.07193}, 2023.

\bibitem[Pan et~al.(2024)Pan, Barath, Pollefeys, and Sch\"{o}nberger]{pan2024glomap}
Linfei Pan, Daniel Barath, Marc Pollefeys, and Johannes~Lutz Sch\"{o}nberger.
\newblock {Global Structure-from-Motion Revisited}.
\newblock In \emph{ECCV}, 2024.

\bibitem[Reizenstein et~al.(2021)Reizenstein, Shapovalov, Henzler, Sbordone, Labatut, and Novotny]{reizenstein21co3d}
Jeremy Reizenstein, Roman Shapovalov, Philipp Henzler, Luca Sbordone, Patrick Labatut, and David Novotny.
\newblock Common objects in 3d: Large-scale learning and evaluation of real-life 3d category reconstruction.
\newblock In \emph{ICCV}, 2021.

\bibitem[Roberts et~al.(2021)Roberts, Ramapuram, Ranjan, Kumar, Bautista, Paczan, Webb, and Susskind]{hypersim}
Mike Roberts, Jason Ramapuram, Anurag Ranjan, Atulit Kumar, Miguel~Angel Bautista, Nathan Paczan, Russ Webb, and Joshua~M. Susskind.
\newblock {Hypersim}: {A} photorealistic synthetic dataset for holistic indoor scene understanding.
\newblock In \emph{ICCV}, 2021.

\bibitem[Rublee et~al.(2011)Rublee, Rabaud, Konolige, and Bradski]{rublee2011orb}
Ethan Rublee, Vincent Rabaud, Kurt Konolige, and Gary Bradski.
\newblock Orb: An efficient alternative to sift or surf.
\newblock In \emph{ICCV}, 2011.

\bibitem[Sarlin et~al.(2020)Sarlin, DeTone, Malisiewicz, and Rabinovich]{sarlin2020superglue}
Paul-Edouard Sarlin, Daniel DeTone, Tomasz Malisiewicz, and Andrew Rabinovich.
\newblock Superglue: Learning feature matching with graph neural networks.
\newblock In \emph{CVPR}, pages 4938--4947, 2020.

\bibitem[Sch\"{o}nberger and Frahm(2016)]{schoenberger2016sfm}
Johannes~Lutz Sch\"{o}nberger and Jan-Michael Frahm.
\newblock Structure-from-motion revisited.
\newblock In \emph{CVPR}, 2016.

\bibitem[Sch\"{o}nberger et~al.(2016)Sch\"{o}nberger, Zheng, Pollefeys, and Frahm]{schoenberger2016mvs}
Johannes~Lutz Sch\"{o}nberger, Enliang Zheng, Marc Pollefeys, and Jan-Michael Frahm.
\newblock Pixelwise view selection for unstructured multi-view stereo.
\newblock In \emph{ECCV}, 2016.

\bibitem[Schops et~al.(2017)Schops, Schonberger, Galliani, Sattler, Schindler, Pollefeys, and Geiger]{eth3d}
Thomas Schops, Johannes~L Schonberger, Silvano Galliani, Torsten Sattler, Konrad Schindler, Marc Pollefeys, and Andreas Geiger.
\newblock A multi-view stereo benchmark with high-resolution images and multi-camera videos.
\newblock In \emph{CVPR}, pages 3260--3269, 2017.

\bibitem[Shen et~al.(2025)Shen, Zhang, Qu, and Cao]{shen2025fastvggt}
You Shen, Zhipeng Zhang, Yansong Qu, and Liujuan Cao.
\newblock Fastvggt: Training-free acceleration of visual geometry transformer.
\newblock \emph{arXiv preprint arXiv:2509.02560}, 2025.

\bibitem[Tang et~al.(2024)Tang, Zhao, Zhu, Xiao, Kasikci, and Han]{tang2024quest}
Jiaming Tang, Yilong Zhao, Kan Zhu, Guangxuan Xiao, Baris Kasikci, and Song Han.
\newblock Quest: Query-aware sparsity for efficient long-context llm inference.
\newblock \emph{ICML}, 2024.

\bibitem[Teed and Deng(2021)]{teed2021droid}
Zachary Teed and Jia Deng.
\newblock Droid-slam: Deep visual slam for monocular, stereo, and rgb-d cameras.
\newblock \emph{NeurIPS}, 2021.

\bibitem[Tillet et~al.(2019)Tillet, Kung, and Cox]{tillet2019triton}
Philippe Tillet, Hsiang-Tsung Kung, and David Cox.
\newblock Triton: an intermediate language and compiler for tiled neural network computations.
\newblock In \emph{Proceedings of the 3rd ACM SIGPLAN International Workshop on Machine Learning and Programming Languages}, 2019.

\bibitem[Wang et~al.(2025{\natexlab{a}})Wang, Schmidt, Piekenbrinck, and Leibe]{wang2025faster}
Chung-Shien~Brian Wang, Christian Schmidt, Jens Piekenbrinck, and Bastian Leibe.
\newblock Faster vggt with block-sparse global attention.
\newblock \emph{arXiv preprint arXiv:2509.07120}, 2025{\natexlab{a}}.

\bibitem[Wang and Agapito(2024)]{wang20243d}
Hengyi Wang and Lourdes Agapito.
\newblock 3d reconstruction with spatial memory.
\newblock \emph{arXiv preprint arXiv:2408.16061}, 2024.

\bibitem[Wang et~al.(2024{\natexlab{a}})Wang, Karaev, Rupprecht, and Novotny]{wang2024vggsfm}
Jianyuan Wang, Nikita Karaev, Christian Rupprecht, and David Novotny.
\newblock Vggsfm: Visual geometry grounded deep structure from motion.
\newblock In \emph{CVPR}, 2024{\natexlab{a}}.

\bibitem[Wang et~al.(2025{\natexlab{b}})Wang, Chen, Karaev, Vedaldi, Rupprecht, and Novotny]{wang2025vggt}
Jianyuan Wang, Minghao Chen, Nikita Karaev, Andrea Vedaldi, Christian Rupprecht, and David Novotny.
\newblock Vggt: Visual geometry grounded transformer.
\newblock In \emph{CVPR}, 2025{\natexlab{b}}.

\bibitem[Wang et~al.(2025{\natexlab{c}})Wang, Zhang, Holynski, Efros, and Kanazawa]{cut3r}
Qianqian Wang, Yifei Zhang, Aleksander Holynski, Alexei~A. Efros, and Angjoo Kanazawa.
\newblock Continuous 3d perception model with persistent state, 2025{\natexlab{c}}.

\bibitem[Wang et~al.(2024{\natexlab{b}})Wang, Xu, Dai, Xiang, Deng, Tong, and Yang]{wang2024moge}
Ruicheng Wang, Sicheng Xu, Cassie Dai, Jianfeng Xiang, Yu Deng, Xin Tong, and Jiaolong Yang.
\newblock Moge: Unlocking accurate monocular geometry estimation for open-domain images with optimal training supervision.
\newblock \emph{arXiv preprint arXiv:2410.19115}, 2024{\natexlab{b}}.

\bibitem[Wang et~al.(2024{\natexlab{c}})Wang, Leroy, Cabon, Chidlovskii, and Revaud]{wang2024dust3r}
Shuzhe Wang, Vincent Leroy, Yohann Cabon, Boris Chidlovskii, and Jerome Revaud.
\newblock Dust3r: Geometric 3d vision made easy.
\newblock In \emph{CVPR}, 2024{\natexlab{c}}.

\bibitem[Wang et~al.(2025{\natexlab{d}})Wang, Zhou, Zhu, Chang, Zhou, Li, Chen, Pang, Shen, and He]{wang2025pi3}
Yifan Wang, Jianjun Zhou, Haoyi Zhu, Wenzheng Chang, Yang Zhou, Zizun Li, Junyi Chen, Jiangmiao Pang, Chunhua Shen, and Tong He.
\newblock $\pi^3$: Scalable permutation-equivariant visual geometry learning.
\newblock \emph{arXiv preprint 2507.13347}, 2025{\natexlab{d}}.

\bibitem[Wu et~al.(2025)Wu, Lin, Zhang, Zeng, Yang, Bao, Qian, Zhu, Torr, Cao, and Yao]{wu2025direct3ds2}
Shuang Wu, Youtian Lin, Feihu Zhang, Yifei Zeng, Yikang Yang, Yajie Bao, Jiachen Qian, Siyu Zhu, Philip Torr, Xun Cao, and Yao Yao.
\newblock Direct3d-s2: Gigascale 3d generation made easy with spatial sparse attention.
\newblock \emph{NeurIPS}, 2025.

\bibitem[Xia et~al.(2024)Xia, Fu, Liu, and Wang]{xia2024rgbd}
Hongchi Xia, Yang Fu, Sifei Liu, and Xiaolong Wang.
\newblock Rgbd objects in the wild: Scaling real-world 3d object learning from rgb-d videos, 2024.

\bibitem[Xiao et~al.(2024)Xiao, Tian, Chen, Han, and Lewis]{xiao2023streamingllm}
Guangxuan Xiao, Yuandong Tian, Beidi Chen, Song Han, and Mike Lewis.
\newblock Efficient streaming language models with attention sinks.
\newblock \emph{ICLR}, 2024.

\bibitem[Yang et~al.(2025)Yang, Sax, Liang, Henaff, Tang, Cao, Chai, Meier, and Feiszli]{yang2025fast3r}
Jianing Yang, Alexander Sax, Kevin~J Liang, Mikael Henaff, Hao Tang, Ang Cao, Joyce Chai, Franziska Meier, and Matt Feiszli.
\newblock Fast3r: Towards 3d reconstruction of 1000+ images in one forward pass.
\newblock \emph{arXiv preprint arXiv:2501.13928}, 2025.

\bibitem[Yao et~al.(2018)Yao, Luo, Li, Fang, and Quan]{yao2018mvsnet}
Yao Yao, Zixin Luo, Shiwei Li, Tian Fang, and Long Quan.
\newblock Mvsnet: Depth inference for unstructured multi-view stereo.
\newblock In \emph{ECCV}, 2018.

\bibitem[Yeshwanth et~al.(2023)Yeshwanth, Liu, Nie{\ss}ner, and Dai]{yeshwanthliu2023scannetpp}
Chandan Yeshwanth, Yueh-Cheng Liu, Matthias Nie{\ss}ner, and Angela Dai.
\newblock Scannet++: A high-fidelity dataset of 3d indoor scenes.
\newblock In \emph{ICCV}, 2023.

\bibitem[Yuan et~al.(2025)Yuan, Gao, Dai, Luo, Zhao, Zhang, Xie, Wei, Wang, Xiao, et~al.]{yuan2025native}
Jingyang Yuan, Huazuo Gao, Damai Dai, Junyu Luo, Liang Zhao, Zhengyan Zhang, Zhenda Xie, Yuxing Wei, Lean Wang, Zhiping Xiao, et~al.
\newblock Native sparse attention: Hardware-aligned and natively trainable sparse attention.
\newblock In \emph{ACL}, 2025.

\bibitem[Zhang et~al.(2025{\natexlab{a}})Zhang, Xiang, Huang, Wei, Xi, Zhu, and Chen]{zhang2025spargeattn}
Jintao Zhang, Chendong Xiang, Haofeng Huang, Jia Wei, Haocheng Xi, Jun Zhu, and Jianfei Chen.
\newblock Spargeattn: Accurate sparse attention accelerating any model inference.
\newblock In \emph{ICML}, 2025{\natexlab{a}}.

\bibitem[Zhang et~al.(2025{\natexlab{b}})Zhang, Chen, Huang, Lin, Liu, Stoica, Xing, and Zhang]{zhang2025vsa}
Peiyuan Zhang, Yongqi Chen, Haofeng Huang, Will Lin, Zhengzhong Liu, Ion Stoica, Eric Xing, and Hao Zhang.
\newblock Vsa: Faster video diffusion with trainable sparse attention.
\newblock \emph{arXiv preprint arXiv:2505.13389}, 2025{\natexlab{b}}.

\bibitem[Zhang et~al.(2023{\natexlab{a}})Zhang, Peng, Hu, and Wang]{geomvsnet}
Zhe Zhang, Rui Peng, Yuxi Hu, and Ronggang Wang.
\newblock Geomvsnet: Learning multi-view stereo with geometry perception.
\newblock In \emph{CVPR}, 2023{\natexlab{a}}.

\bibitem[Zhang et~al.(2023{\natexlab{b}})Zhang, Sheng, Zhou, Chen, Zheng, Cai, Song, Tian, R{\'e}, Barrett, et~al.]{zhang2023h2o}
Zhenyu Zhang, Ying Sheng, Tianyi Zhou, Tianlong Chen, Lianmin Zheng, Ruisi Cai, Zhao Song, Yuandong Tian, Christopher R{\'e}, Clark Barrett, et~al.
\newblock H2o: Heavy-hitter oracle for efficient generative inference of large language models.
\newblock \emph{NeurIPS}, 2023{\natexlab{b}}.

\bibitem[Zhou et~al.(2018)Zhou, Tucker, Flynn, Fyffe, and Snavely]{zhou2018stereo}
Tinghui Zhou, Richard Tucker, John Flynn, Graham Fyffe, and Noah Snavely.
\newblock Stereo magnification: learning view synthesis using multiplane images.
\newblock \emph{ACM TOG}, 2018.

\end{thebibliography}
}
\clearpage
\setcounter{page}{1}
\setcounter{section}{0}

\maketitlesupplementary

\section{Efficient Kernel Implementation}
\label{sec:supp_kernel}
\subsection{Implementation Details}
Our Gated Sparse Attention (GSA) utilizes a dedicated kernel for each branch. For the compression branch, we adapt the official FlashAttention implementation in Triton by integrating a streaming top-k selection, which employs a bitonic search algorithm, directly within the online softmax computation. The pseudocode for this forward kernel is presented in Algorithm~\ref{alg:gsa_forward_detailed}.

For the selection kernel, it shares a similar idea with block-sparse attention of original Flash Attention2. But add custom api to get topk-idx from the compression kernel and mask the first frame token for vggt, as shown in Algorithm ~\ref{alg:blocksparse_forward}. We do not show the pseudocode of the backward kernel but we implement them following originla NSA~\cite{yuan2025native} and FlashAttention2~\cite{dao2023flashattention2} logic.

We observe that the two branches of our proposed method are conceptually analogous to two recent training-free approaches.

The philosophy behind our compression branch is shared by the token merging strategy in FastVGGT~\cite{shen2025fastvggt}. Both methods aim to reduce sequence length by merging redundant tokens to improve efficiency. However, they differ in their implementation: FastVGGT employs a bipartite matching algorithm for merging, whereas our approach utilizes a more efficient token average pooling mechanism.

Similarly, our selection branch aligns with the strategy used in Block-Sparse attention~\cite{wang2025faster}, which also selects top-k key blocks for each query to reduce computational overhead. A limitation of their work is the reliance on a pre-existing kernel~\cite{zhang2025spargeattn} that lacks an implementation for the backward pass. This makes their method non-differentiable and thus non-trainable, hindering its scalability and practical utility. In contrast, our method introduces a custom, fully differentiable kernel, enabling end-to-end training.

To support the broader research community, we will open-source our kernel implementation, offering a scalable and trainable solution for efficient attention.

\begin{algorithm}[htbp]
\caption{Compression Branch Forward Pass}
\label{alg:gsa_forward_detailed}
\begin{algorithmic}[1]
\Require Query $Q$, Key $K$, Value $V$
\Require Top-k value $k_{\text{top}}$, Block dimensions $B_M, B_N$, Head dimension $d_h$
\Ensure Output $O$, Top-k indices $I_{\text{topk}}$

\Procedure{GSA\_Forward\_Detailed}{$Q, K, V, k_{\text{top}}$}
    \For{each query block $Q_i$ of size $B_M \times d_h$ \textbf{in parallel}}
        \State Initialize output accumulator $O_i \leftarrow \mathbf{0}$
        \State Initialize row-wise max statistic $m_i \leftarrow -\infty$
        \State Initialize row-wise log-sum-exp statistic $l_i \leftarrow \mathbf{0}$
        \State Initialize top-k scores and indices: $(\mathcal{S}_i, \mathcal{I}_i) \leftarrow (-\infty, \text{null})$

        \For{each key/value block $(K_j, V_j)$ of size $B_N \times d_h$}
            \State Compute attention scores $S_{ij} \leftarrow (Q_i K_j^T)$

            \Statex \Comment{1. Update attention output via online softmax}
            \State $m_{ij} \leftarrow \rowmax(S_{ij})$
            \State $P_{ij} \leftarrow \exp(S_{ij} - m_{ij})$ 
            \State $l_{ij} \leftarrow \rowsum(P_{ij})$
            \State $m_{i}^{\text{new}} \leftarrow \max(m_i, m_{ij})$
            \State $l_{i}^{\text{new}} \leftarrow e^{m_i - m_{i}^{\text{new}}} l_i + e^{m_{ij} - m_{i}^{\text{new}}} l_{ij}$
            \State $O_i \leftarrow e^{m_i - m_{i}^{\text{new}}} O_i + e^{m_{ij} - m_{i}^{\text{new}}} (P_{ij} V_j)$
            \State $m_i \leftarrow m_{i}^{\text{new}}$; $l_i \leftarrow l_{i}^{\text{new}}$

            \Statex \Comment{2. Update top-k indices via streaming selection}
            \State $(S_{ij}^{\text{top}}, I_{ij}^{\text{top}}) \leftarrow \TopK(S_{ij}, k_{\text{top}})$ 
            \State $(\mathcal{S}_i, \mathcal{I}_i) \leftarrow \text{Sort}((\mathcal{S}_i, \mathcal{I}_i), (S_{ij}^{\text{top}}, I_{ij}^{\text{top}}), k_{\text{top}})$
        \EndFor

        \State Normalize final block output $O_i \leftarrow O_i / l_i$
        \State Store block output $O_i$ into $O$ and final top-k indices $\mathcal{I}_i$ into $I_{\text{topk}}$.
    \EndFor
\EndProcedure
\end{algorithmic}
\end{algorithm}

\begin{algorithm}[htbp]
\caption{Selection Attention Forward Pass}
\label{alg:blocksparse_forward}
\begin{algorithmic}[1]
\Require Query $Q$, Key $K$, Value $V$, Top-k indices tensor $T$
\Require Softmax scale $\sigma$, Block dimensions $B_M, B_N$, Head dimension $d_h$, Top-k count $k_{\text{top}}$
\Ensure Output $O$, Log-Sum-Exp values $L$

\Procedure{BlockSparse\_Forward}{}
    \For{each query block $Q_i$ of size $B_M \times d_h$ \textbf{in parallel}}
        \State Initialize output accumulator $O_i \leftarrow \mathbf{0}$
        \State Initialize row-wise max statistic $m_i \leftarrow -\infty$
        \State Initialize row-wise sum-exp statistic $l_i \leftarrow \mathbf{0}$

        \For{$k$ from $0$ to $k_{\text{top}}-1$}
            \State Load key block index $j \leftarrow T_{i,k}$
            \State Load corresponding key block $K_j$ and value block $V_j$
            
            \State Compute attention scores $S_{ij} \leftarrow \sigma (Q_i K_j^T)$
            
            \Statex \Comment{Update attention output via online softmax}
            \State $m_{ij} \leftarrow \rowmax(S_{ij})$
            \State $P_{ij} \leftarrow \exp(S_{ij} - m_{ij})$ 
            \State $l_{ij} \leftarrow \rowsum(P_{ij})$
            \State $m_{i}^{\text{new}} \leftarrow \max(m_i, m_{ij})$
            \State $l_{i}^{\text{new}} \leftarrow e^{m_i - m_{i}^{\text{new}}} l_i + e^{m_{ij} - m_{i}^{\text{new}}} l_{ij}$
            \State $O_i \leftarrow e^{m_i - m_{i}^{\text{new}}} O_i + e^{m_{ij} - m_{i}^{\text{new}}} (P_{ij} V_j)$
            \State $m_i \leftarrow m_{i}^{\text{new}}$; $l_i \leftarrow l_{i}^{\text{new}}$
        \EndFor

        \State Compute final Log-Sum-Exp $L_i \leftarrow m_i + \log(l_i)$
        \State Normalize final block output $O_i \leftarrow O_i / l_i$
        
        \State Store block output $O_i$ into $O$ and statistics $L_i$ into $L$.
    \EndFor
\EndProcedure
\end{algorithmic}
\end{algorithm}

\section{Speed3R-VGGT Benchmarking}
\label{sec:speed3r-vggt-bench}
\begin{table}[t] 
\caption{\textbf{Inference Time for Different Models.} Mean latency (in seconds) for varying sequence lengths on VGGT backbone. Our method achieves a 10.9× speedup on sequences of 1024 images.}
\label{tab:latency_vggt}
\centering
\noindent\resizebox{\columnwidth}{!}{%
\begin{tabular}{lcccccc}
\toprule
\textbf{Seq Length} & \textbf{32} & \textbf{64} & \textbf{128} & \textbf{256} & \textbf{512} & \textbf{1024} \\
\midrule
Full Attn.(VGGT) & 0.79 & 1.81 & 5.44 & 18.21 & 67.27 & 271.21 \\
FastVGGT~\cite{wang2025faster} & 0.63 & 1.25 & 2.59 & 6.48 & 18.30 & 61.33 \\
Block Sparse~\cite{shen2025fastvggt} & 0.53 & 1.16 & 2.89 & 8.16 & 26.15 & 75.42 \\
Ours & 0.46 & 0.87 & 1.78 & 3.88 & 9.16 & 24.85 \\
\bottomrule
\end{tabular}%
}
\end{table}

We benchmark the inference acceleration ratio of Speed3R-VGGT, as shown in Table~\ref{tab:latency_vggt}. Our method achieves a 10.9× speedup compared to full attention and demonstrates greater efficiency than training-free methods. The latency is measured in pose-only mode. Note that in all implementations, we utilize VGGT with redundant memory storage removed, as proposed in FastVGGT~\cite{shen2025fastvggt}, to benchmark the latency of VGGT. This optimization achieves an acceleration ratio of approximately 2 compared to the vanilla VGGT implementation.

\section{Training/Evaluation Details}
\label{sec:details}
\begin{table}[h]
\centering
\caption{Summary of Datasets Used in Training and Testing}
\label{tab:dataset_summary}
\resizebox{\columnwidth}{!}{%
\begin{tabular}{@{}lll@{}}
\toprule
\textbf{Dataset Name} & \textbf{Scene/Sequence Count} & \textbf{Downsample Ratio} \\
\midrule
Co3Dv2~\cite{reizenstein21co3d}   & 10764 & 2x \\
Hypersim~\cite{hypersim}   & 461 & 1x \\
Scannetpp~\cite{yeshwanthliu2023scannetpp}   & 996 & 2x \\
DL3DV~\cite{ling2024dl3dv}   & 6,378 & 4x \\
ARKitScenes~\cite{dehghan2021arkitscenes}   & 3,344 & 2x \\
WildRGBD~\cite{xia2024rgbd}   & 2,753 & 5x\\
vkitti~\cite{cabon2020vkitti2}   & 5  & 1x \\
\bottomrule
\end{tabular}%
}
\end{table}

\paragraph{Datasets} The statistics of our training datasets are detailed in Table~\ref{tab:dataset_summary}. To manage storage requirements, we downsampled each sequence at the ratios specified in the table. During training, we sample 2-24 images following previous works. For the sampling process, we employed random sampling for most datasets. However, for the VKITTI2 dataset, which is characterized by its long sequences, we adopted the sampling strategy from $\pi^3$~\cite{wang2025pi3}. The total size of the processed training data amounts to approximately 2TB, excluding depth maps. Our distillation-based strategy obviates the need to store these large depth files, resulting in significant storage savings.

\paragraph{Speed3R-VGGT Training Loss} Due to the self-normalization nature of VGGT, we do not need to normalize prediction and teacher supervision.The Speed3R-VGGT is trained end-to-end with a composition of two terms: $\mathcal{L}_{\text{camera}}$, $\mathcal{L}_{\text{depth}}$. The camera loss, $\mathcal{L}_{\text{camera}}$, supervises a set of $N$ camera pose predictions ($\hat{g}i$) by comparing them against their teacher model prediction ($g_i$) using the Huber loss. The depth loss, $\mathcal{L}{\text{depth}}$, implementing an aleatoric uncertainty loss that leverages a predicted uncertainty map, $\Sigma_i^D$, which acts as a measure of confidence. This uncertainty map is used to weigh the discrepancy between the predicted depth ($\hat{D}i$) and pseudo depth GT from teacher model($D_i$), and it also weighs an additional gradient-based term ($\nabla\hat{D}i - \nabla D_i$) to preserve structural details. The depth loss formulation, $\mathcal{L}{\text{depth}} = |\Sigma_i^D \odot (\hat{D}_i - D_i)| + |\Sigma_i^D \odot (\nabla\hat{D}_i - \nabla D_i)| - \alpha \log \Sigma_i^D$, therefore penalizes depth and gradient errors based on the model's own confidence while also regularizing the uncertainty prediction itself.

\paragraph{Speed3R-$\mathbf{\pi^3}$ Training Loss}
Analogous to the training of Speed3R-VGGT, the loss function for Speed3R-$\pi^3$ is a composite objective comprising two primary components: one for the depth head and another for the pose head. A key aspect of this loss, adopted from the $\pi^3$ methodology~\cite{wang2025pi3}, is the normalization of both the model's predictions and the teacher supervision signals by the average depth value computed across all views. The overall loss function adheres to the original $\pi^3$ implementation, incorporating several specific terms: an aligned depth loss, a surface normal loss derived from MoGe~\cite{wang2024moge}, and a relative pose loss that is computed on all pairs of predicted camera poses.

\paragraph{Train from Scratch}
In section~\ref{sec:discussion}, we train our sparse model and dense model from scratch. We train both model with 8GPU for 40 epoch with gradient accumulation of 2. We initialize two model encoder with DINOv2~\cite{oquab2023dinov2} weights and other part with Kaiming Initialization~\cite{he2015delving}.
\section{Sparsity Calculation}

\label{sec:Sparsity}
For our method and Block-Sparse~\cite{wang2025faster}, we report the sparse ratio based on the selection stage, as both the query (q) and key (k) are downsampled by a factor of 16 in the compression stage. This results in a very small sparse ratio in terms of FLOPs.

For FastVGGT~\cite{shen2025fastvggt}, we report their threshold-based sparse ratio. Their method sparsifies both the query (q) and key (k), which helps reduce FLOPs during the attention calculation. However, it introduces an additional pre-filtering stage, where the maximum similarity score of each token is computed relative to the protected tokens. This step adds quadratic complexity to their method.

Additionally, FastVGGT sets the first view as protected tokens that cannot be evicted. As a result, when the sparse ratio is set to 0.9, the actual effective sparse ratio is approximately 0.81. From a FLOPs perspective, we report their sparse ratio while accounting for this behavior and the overhead introduced by the pre-filtering stage.


\end{document}